\documentclass[12pt,fleqn]{wlscirep}
\usepackage{amsmath}
\usepackage{amssymb}
\usepackage[utf8]{inputenc}
\usepackage[T1]{fontenc}
\usepackage{titling}
\usepackage{bm}
\usepackage{setspace}
\usepackage{caption}
\usepackage{graphicx}    
\usepackage{subcaption}  
\usepackage{multirow}
\usepackage{float}
\usepackage{booktabs}
\usepackage{tabularx}
\usepackage{array}
\usepackage{longtable}
\usepackage{parskip}
\usepackage{soul}
\usepackage{orcidlink}

\usepackage{xcolor}
\usepackage{tikz}
\usetikzlibrary{positioning,fit,calc,arrows.meta}

\begin{document}

\begin{titlepage}
\raggedright
\textbf{Review Article} \\
\vspace{1.0cm}

\begin{center}

\LARGE
\textbf{Agentic AI in medicine: architectures, applications, evaluation, and challenges for clinical translation}
\\

\vspace{1.2cm}

\small
\textbf{
Wanshu Fan\textsuperscript{1}~\orcidlink{0000-0001-6299-2795},
Zheng Tong\textsuperscript{1},
Yang Liu\textsuperscript{1},
Jing Qin\textsuperscript{1,*},
Zhongbin Han\textsuperscript{2},
Haifan Gong\textsuperscript{3},
Congyu Liao\textsuperscript{4},
Xiaofeng Liu\textsuperscript{5},
Cong Wang\textsuperscript{6,*}~\orcidlink{0000-0002-6068-0103}} \\
\vspace{0.3cm}

\textsuperscript{1} National and Local Joint Engineering Laboratory of Computer Aided Design, School of Software Engineering, Dalian University, Dalian, China\\
\textsuperscript{2}Department of Surgery, Affiliated Zhongshan Hospital of Dalian University, Dalian, China\\
\textsuperscript{3}The Chinese University of Hong Kong, Shenzhen, China\\
\textsuperscript{4}University of California, San Francisco, USA\\
\textsuperscript{5}Yale University, New Haven, USA\\
\textsuperscript{6}The Hong Kong Polytechnic University, Hong Kong, China
\end{center}

\vspace{0.3cm}
{Correspondence and requests for materials should be addressed to:\\
    Jing Qin: qinjing@dlu.edu.cn\\
Cong Wang: supercong94@gmail.com}



\vfill
\end{titlepage}

\newpage
\raggedright
\textbf{\large Abstract}

\vspace{1cm}
%


Large language models and multimodal foundation models are enabling medical
artificial intelligence (AI) systems to move beyond isolated prediction and undertake multistep clinical tasks that require planning, tool use, memory, iterative correction, and coordination among specialized agents. 
However, the scope of agentic AI in medicine remains unsettled, and current evaluation practices are not yet aligned with the requirements of clinical use. 
We conducted a scoping review with systematic evidence mapping across five electronic sources, screened 1,649 exportable records, and provisionally included 557 unique studies that met predefined criteria for goal-directed task execution, tool use, interaction with external resources, feedback-based refinement, or multi-agent collaboration. The included studies describe single agents that use external tools, workflows supported by retrieval and external knowledge, multimodal agents, and multi-agent systems applied to medical question answering, image interpretation, electronic health record analysis, drug safety, and clinical trial prediction. 
The evidence base remains dominated by public benchmarks, simulated settings, retrospective datasets, and small-scale expert evaluation. Process reliability, evidence traceability, uncertainty, safety, workflow impact, and external validity are evaluated less consistently. 
Clinical translation will depend on clearer definitions, reproducible evaluation, auditable oversight, interoperable system design, and prospective validation in real-world clinical workflows.

\vspace{2cm}


\newpage

\setstretch{1.2}

\section{Introduction}

Medical artificial intelligence (AI) has advanced substantially over the past decade, with established applications in medical imaging, risk prediction, and clinical decision support \cite{Topol2019HighPerformanceMedicine,Rajpurkar2022AIHealthMedicine}. Earlier systems were predominantly developed for well-defined tasks, including image classification, lesion detection, segmentation, risk estimation, and text generation. Although effective within predefined settings, these models generally map inputs directly to task-specific outputs and provide limited support for multistep reasoning, adaptive interaction, or explicit coordination of external resources. Large language models (LLMs), vision--language models, and multimodal foundation models have broadened this paradigm by supporting medical question answering, report generation, clinical knowledge reasoning, and cross-modal information integration \cite{Moor2023GMAI,Singhal2023ClinicalKnowledge,Singhal2025ExpertMedicalQA,Zhou2025LLMBiomedicine,AlSaad2024MultimodalLLMHealth,Tu2025ConversationalDiagnosticAI,Tu2024GeneralistBiomedicalAI,Thirunavukarasu2023LLMMedicine,Clusmann2023FutureLLMMedicine}. These developments provide the representational and generative capabilities required for more general medical AI systems.

%
Foundation models, however, are not equivalent to Agentic AI. Foundation models primarily provide language understanding, image interpretation, knowledge representation, and multimodal reasoning, whereas Agentic AI organizes these capabilities within goal-directed workflows. Such workflows may involve task decomposition, memory, external knowledge retrieval, tool or model invocation, execution monitoring, and feedback-based revision. In medical image analysis, for example, a vision-language model may interpret an image or generate a report, while an agentic system may additionally invoke detection or segmentation models, retrieve relevant electronic health record (EHR) information and clinical guidance, and integrate the resulting evidence into a traceable workflow. General agent frameworks such as ReAct, Toolformer, and AutoGen provide mechanisms for combining reasoning with external actions, tool use, and collaboration among specialized agents \cite{Yao2023ReAct,Schick2023Toolformer,Wu2024AutoGen}. Within medicine, Agentic AI can therefore be viewed as a system-level paradigm for coordinating models, tools, knowledge sources, and intermediate decisions around a defined clinical objective.

This paradigm is particularly relevant to clinical tasks that require the integration of heterogeneous and temporally distributed information. Diagnostic and treatment decisions may depend on medical images, pathology findings, laboratory results, physiological signals, medication records, longitudinal electronic health records, clinical guidelines, and patient-reported information. These sources are often reviewed under conditions of incomplete information, time pressure, and substantial clinical risk. Conventional task-specific models usually address only one component of this process. Agentic systems may extend their utility by selecting appropriate resources, maintaining intermediate task states, revising actions in response to new evidence, and coordinating specialized components across a multistep workflow. Such capabilities do not imply unrestricted clinical autonomy. Instead, they introduce additional requirements for action boundaries, evidence provenance, error recovery, human oversight, and accountability.

The field nevertheless remains conceptually fragmented and methodologically heterogeneous. The terms AI agent, LLM-based agent, and Agentic AI are used inconsistently, with studies placing different emphasis on autonomy, goal-directed behavior, tool use, memory, feedback, and multi-agent collaboration. Proposed architectures range from single agents that invoke external tools to multi-agent systems that emulate multidisciplinary consultation. Other systems operate as workflow agents integrated with electronic health records, picture archiving and communication systems, Fast Healthcare Interoperability Resources (FHIR), clinical guidelines, or drug databases. Evaluation frameworks have not kept pace with architectural development. Most studies continue to emphasize task-level metrics such as accuracy, area under the receiver operating characteristic curve, and F1-score, whereas tool selection, execution reliability, evidence traceability, error recovery, interaction safety, clinician oversight, and workflow impact are assessed less consistently. Clinical evidence also remains limited, as many systems are evaluated using public benchmarks, simulated environments, retrospective datasets, or small-scale expert assessment rather than prospective studies or routine clinical workflows.

To address these limitations, this scoping review systematically examines the conceptual foundations, technical architectures, clinical applications, evaluation practices, and translation readiness of Agentic AI in medicine. We clarify the distinctions among AI agents, LLM-based agents, and Agentic AI; characterize their core capabilities and technical architectures; map applications across medical domains; and assess current evaluation practices and levels of clinical validation. The analysis is organized around six dimensions: task performance, process reliability, evidence traceability, safety, clinical utility, and external validation. By distinguishing demonstrated technical capability from evidence of clinical value, the review identifies the methodological and practical requirements for reproducible, auditable, and responsible clinical translation.

Surveys, reviews, and perspectives on AI agents have accumulated rapidly. General surveys have characterized the construction, capabilities, and evaluation of LLM-based autonomous agents and multi-agent systems \cite{Wang2024LLMAgentsSurvey,Xi2025RiseOfAgents,Guo2024MultiAgentsSurvey,Sumers2024CoALA}, whereas medicine-focused perspectives have outlined the opportunities and risks of agentic systems for clinical care, biomedical discovery, and health-system coordination \cite{Qiu2024AgenticSystemsMedicine,Gao2024BiomedicalAgents,Mehandru2024AgentsInClinic,Moritz2025CoordinatedAgents,Zou2025AgenticTeammates}. More recent reviews have begun to synthesize healthcare-specific agent studies, taxonomies, and evaluation practices \cite{Wang2025BaymaxSurvey,Collaco2026AgenticHealthcare,Zhao2026AIAgentHealthcare,Vatsal2026AgenticTaxonomy}. Table~\ref{tab:prior_reviews} summarizes how these works differ in type, scope, and emphasis. In contrast to these works, the present review combines a predefined operational definition of medical Agentic AI with systematic evidence mapping and explicitly separates functional sophistication from the maturity of clinical validation.

\begin{table}[!htbp]
\centering
\caption{Comparison of this review with representative surveys, reviews, and perspectives on LLM-based agents and Agentic AI. General surveys provide technical foundations but are not specific to medicine, medicine-focused perspectives articulate opportunities and risks, and healthcare reviews differ in definitional breadth and analytical emphasis. Descriptions summarize the stated scope of each work and do not constitute a quality assessment.}
\label{tab:prior_reviews}
\footnotesize
\renewcommand{\arraystretch}{1.15}
\begin{tabularx}{\textwidth}{
>{\raggedright\arraybackslash}p{3.0cm}
>{\raggedright\arraybackslash}p{2.4cm}
>{\raggedright\arraybackslash}p{3.0cm}
>{\raggedright\arraybackslash}X}
\toprule
Work & Type & Scope & Principal emphasis \\
\midrule
Wang et al.~\cite{Wang2024LLMAgentsSurvey} & Survey & General LLM-based autonomous agents & Agent construction (profile, memory, planning, and action), applications, and evaluation \\
Xi et al.~\cite{Xi2025RiseOfAgents} & Survey & General LLM-based agents & Brain--perception--action framework, single-agent and multi-agent settings, and agent societies \\
Guo et al.~\cite{Guo2024MultiAgentsSurvey} & Survey & LLM-based multi-agent systems & Agent profiling, communication, and capability acquisition \\
Sumers et al.~\cite{Sumers2024CoALA} & Conceptual framework & Language agents & Modular memory, structured action spaces, and decision procedures \\
Qiu et al.~\cite{Qiu2024AgenticSystemsMedicine} & Perspective & Medicine and healthcare & Opportunities and risks of LLM-based agentic systems \\
Gao et al.~\cite{Gao2024BiomedicalAgents} & Perspective & Biomedical discovery & AI agents as research collaborators and levels of autonomy \\
Mehandru et al.~\cite{Mehandru2024AgentsInClinic} & Perspective & Clinical evaluation & Evaluation of LLM agents in simulated clinical environments \\
Moritz et al.~\cite{Moritz2025CoordinatedAgents} & Perspective & Healthcare delivery & Coordination among multiple specialized AI agents \\
Zou and Topol~\cite{Zou2025AgenticTeammates} & Comment & Clinical care & Agentic AI systems as clinical teammates \\
Wang et al.~\cite{Wang2025BaymaxSurvey} & Survey & Medicine & Architectures, applications, and evaluation of LLM-based medical agents \\
Collaco et al.~\cite{Collaco2026AgenticHealthcare} & Scoping review & Healthcare & Restrictive definition of Agentic AI applied to a small set of eligible studies \\
Zhao et al.~\cite{Zhao2026AIAgentHealthcare} & Review & Healthcare & Broad AI-agent applications, evaluations, and future directions \\
Vatsal et al.~\cite{Vatsal2026AgenticTaxonomy} & Taxonomy & Healthcare and medicine & Multidimensional taxonomy for empirical characterization of agentic systems \\
This review & Scoping review with evidence mapping & Medicine (557 studies) & Operational definition, architectures, applications, evaluation, and validation maturity \\
\bottomrule
\end{tabularx}
\end{table}

The main contributions of this review are fourfold. First, it proposes an operational distinction among AI agents, LLM-based agents, and Agentic AI systems that is grounded in implemented mechanisms rather than terminology. Second, it organizes existing systems into recurrent architecture families and orchestration patterns and relates them to foundation models, specialized tools, retrieval, memory, and clinical interoperability standards. Third, it maps applications across medical imaging, clinical reasoning, EHR analysis, drug safety, biomedical research, and other clinically relevant workflows such as treatment planning and patient-facing services, together with the datasets, benchmarks, and simulation environments used to evaluate them. Fourth, it proposes a six-dimensional evaluation framework and a staged translation pathway, linked to failure modes, reporting guidelines, and governance requirements, to support reproducible and auditable clinical translation.

The remainder of the review is organized as follows. The Methods section describes the review design, search strategy, eligibility criteria, study selection, data extraction, and evidence appraisal. The Results section summarizes the search results and research landscape; defines the core capabilities and technical architectures of medical Agentic AI; introduces conceptual frameworks for agentic clinical workflows; reviews applications; and examines evaluation and validation practices. The Discussion synthesizes the main findings, compares them with existing reviews, analyzes the limitations of current research and of this review, discusses safety, governance, and clinical translation, and outlines future research directions.

\section{Methods}

\subsection{Review Design}



%
We conducted a scoping review with a systematic evidence-mapping component. The methodological approach was informed by the framework proposed by Arksey and O'Malley, and reporting was guided by the Preferred Reporting Items for Systematic Reviews and Meta-Analyses Extension for Scoping Reviews (PRISMA-ScR) \cite{Arksey2005Scoping,Tricco2018PRISMAScR}. A scoping review was selected because the literature on medical Agentic AI was heterogeneous in terminology, system boundaries, technical architectures, application settings, and evaluation practices. The review aimed to characterize the scope, composition, and maturity of the available evidence rather than to estimate a pooled treatment effect or perform a quantitative meta-analysis.

Systematic evidence mapping was used to organize the included studies across predefined dimensions, including agentic capability, technical architecture, medical application domain, evaluation strategy, and level of clinical validation \cite{Petersen2008Mapping}. This component supported categorical synthesis and visualization of the evidence and was not treated as a separate quantitative systematic review.

\subsection{Search Strategy}

We searched five electronic sources: PubMed, IEEE Xplore, Europe PMC, arXiv, and medRxiv. These sources were selected to capture biomedical and clinical research, computer science and engineering literature, and rapidly emerging preprint studies. The search covered English-language records published between 1 January 2022 and 23 July 2026. Backward reference screening and forward citation tracking of relevant primary studies and reviews were also performed to identify reports not retrieved through the electronic searches.

The search strategy combined agent-related terms, including ``agentic AI'', ``AI agent'', ``LLM-based agent'', ``large language model agent'', ``medical agent'', ``multi-agent system'', ``autonomous agent'', and ``tool-using agent'', with medical and clinical terms, including medicine, healthcare, clinical care, biomedical research, medical imaging, electronic health records, clinical decision support, drug safety, and clinical trials. Search syntax was adapted to the indexing structure and search functions of each source while retaining the same conceptual structure.

Representative general-agent studies, including ReAct, Toolformer, and AutoGen, were retained as technical background because of their influence on tool use, multistep reasoning, and multi-agent collaboration \cite{Yao2023ReAct,Schick2023Toolformer,Wu2024AutoGen}. Earlier medical agent studies were used, where necessary, to describe the progression from rule-based and reinforcement-learning systems to contemporary LLM-based agents. These studies were not included in the systematic evidence mapping unless they met all predefined eligibility criteria.

\subsection{Eligibility Criteria}

Studies were eligible for inclusion in the systematic evidence mapping if they met all of the following criteria: (1) they were English-language primary research reports published between 1 January 2022 and 23 July 2026 as peer-reviewed journal articles, full-length conference papers, or publicly accessible preprints; (2) they investigated a substantive medical, healthcare, or clinically relevant biomedical application; (3) they described or evaluated a goal-directed system that performed a multistep task; and (4) the implemented system incorporated at least one explicit agentic mechanism, including task planning or decomposition, external tool or code use, interaction with an external resource or environment, state or memory maintenance across task steps, feedback-based refinement, or multi-agent collaboration. Eligibility was determined from the implemented workflow and evaluation rather than from the use of the terms ``agent'', ``AI agent'', or ``Agentic AI'' alone.

Studies were excluded if they met any of the following criteria: (1) lack of relevance to medicine, healthcare, or clinically relevant biomedical research; (2) evaluation of a general-purpose agent without a substantive medical application; (3) evaluation of a standalone foundation model, large language model, vision--language model, retrieval system, or conventional single-task medical AI model that lacked goal-directed multistep execution or an explicit agentic mechanism; (4) publication as a review, editorial, commentary, perspective, protocol, thesis, student project report, conference abstract, or poster-only report; or (5) insufficient methodological or evaluation detail to determine the implemented workflow, data source, task setting, or validation procedure.

Duplicate and substantially overlapping reports were consolidated at the study level. When a preprint and a peer-reviewed publication described the same underlying study, the peer-reviewed version was retained. When no peer-reviewed version was available, the most recent accessible preprint was used. Each underlying study was counted once in the systematic evidence mapping. Earlier medical AI studies, general agent-framework papers, datasets, evaluation-metric studies, reporting guidelines, and methodological references were retained as contextual or technical background but were not counted as included studies unless they independently met all eligibility criteria. Accordingly, the reference list of this review is not equivalent to the evidence-mapping set. Agentic systems discussed in the Results are cited as representative examples of recurrent architectures, applications, and evaluation designs, whereas foundation models, general agent frameworks, datasets, evaluation methods, reporting guidelines, and studies of safety, bias, and human factors are cited as contextual or technical background. For studies available both as preprints and as peer-reviewed publications, the version of record is cited.

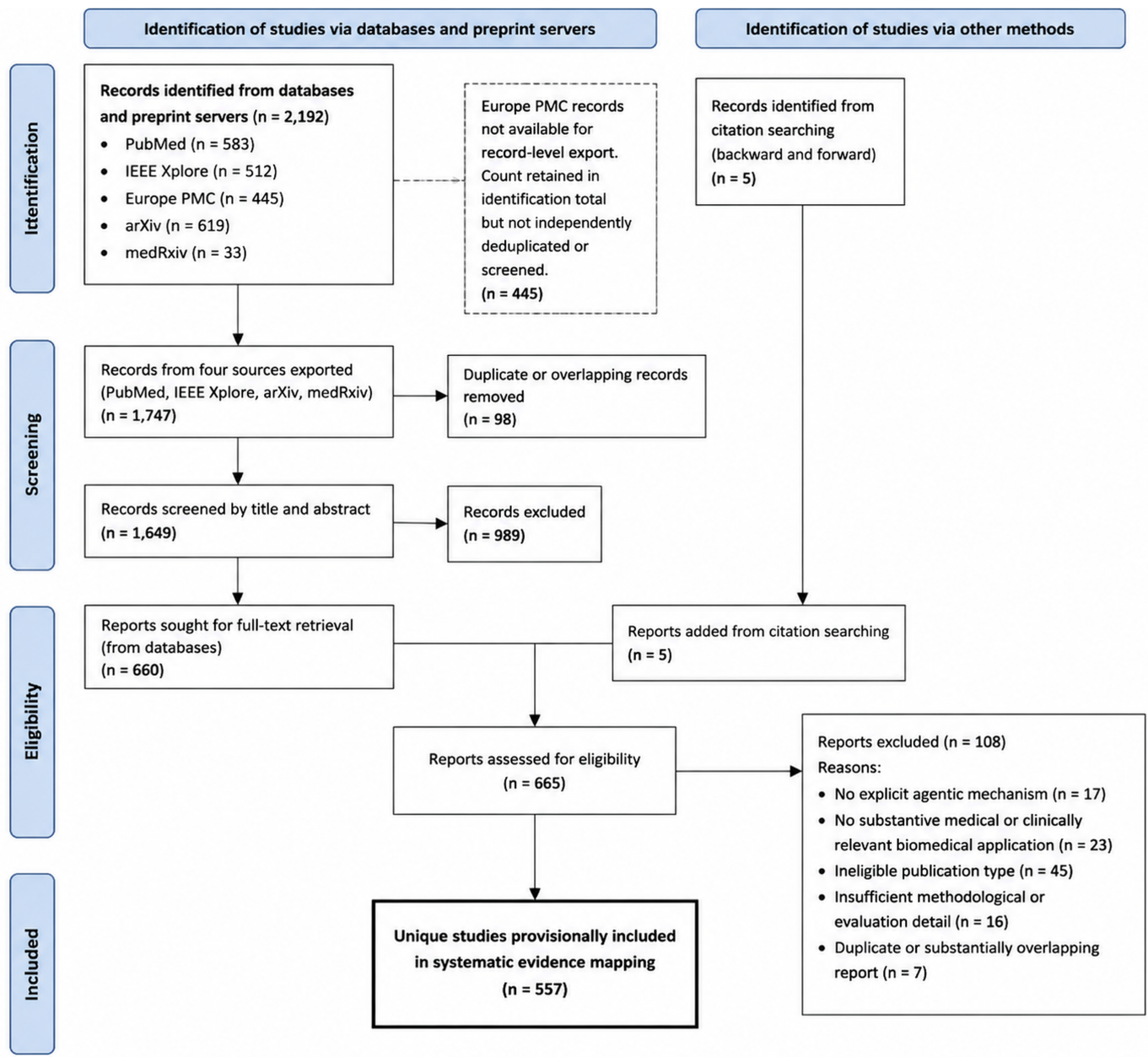
\begin{figure}[!t]
    \centering
\begin{tikzpicture}[
  font=\sffamily\scriptsize,
  every node/.append style={execute at begin node={\hyphenpenalty=10000\exhyphenpenalty=10000}},
  >={Stealth[length=2mm]},
  box/.style={draw, align=left, inner sep=3.5pt, line width=0.5pt},
  head/.style={draw=blue!50!black, fill=blue!8, rounded corners=3pt, align=center, font=\sffamily\scriptsize\bfseries, inner sep=4pt},
  phase/.style={draw=blue!50!black, fill=blue!8, rounded corners=3pt, inner sep=0pt},
  phasetext/.style={rotate=90, font=\sffamily\scriptsize\bfseries},
  arr/.style={->, line width=0.6pt},
  ln/.style={line width=0.6pt}
]
\node[head, minimum width=10.03cm] (h1) at (6.225,0) {Identification of studies via databases and preprint servers};
\node[head, minimum width=4.28cm] (h2) at (14.0,0) {Identification of studies via other methods};
\node[box, text width=4.8cm, anchor=north] (id1) at (3.75,-0.7) {\textbf{Records identified from databases and preprint servers (n~=~2,192)}\\[2pt]
\textbullet~PubMed (n~=~583)\\
\textbullet~IEEE Xplore (n~=~512)\\
\textbullet~Europe PMC (n~=~445)\\
\textbullet~arXiv (n~=~619)\\
\textbullet~medRxiv (n~=~33)};
\node[box, dashed, text width=4.2cm, anchor=north] (epmc) at (9.0,-0.7) {Europe PMC records not available for record-level export; count retained in the identification total but not independently deduplicated or screened (n~=~445)};
\node[box, text width=4.0cm, anchor=north] (oth1) at (14.0,-0.7) {Records identified from citation searching (backward and forward) (n~=~5)};
\node[box, text width=4.8cm, anchor=north] (s1) at ([yshift=-0.75cm]id1.south) {Records from four sources exported (PubMed, IEEE Xplore, arXiv, medRxiv) (n~=~1,747)};
\node[box, text width=4.2cm] (dup) at (s1 -| epmc) {Duplicate or substantially overlapping records removed (n~=~98)};
\node[box, text width=4.8cm, anchor=north] (s2) at ([yshift=-0.55cm]s1.south) {Records screened by title and abstract (n~=~1,649)};
\node[box, text width=4.2cm] (ex1) at (s2 -| epmc) {Records excluded (n~=~989)};
\node[box, text width=4.8cm, anchor=north] (e0) at ([yshift=-0.75cm]s2.south) {Reports sought for full-text retrieval (from databases) (n~=~660)};
\node[box, text width=4.0cm] (oth2) at (e0 -| oth1) {Reports added from citation searching (n~=~5)};
\coordinate (join) at (e0 -| epmc);
\node[box, text width=3.6cm, align=center, anchor=north] (e1) at ([yshift=-0.8cm]join) {Reports assessed for eligibility (n~=~665)};
\node[box, text width=4.0cm, anchor=north] (ex2) at (e1.north -| oth1) {\textbf{Reports excluded (n~=~108)}\\[2pt]
\textbullet~No explicit agentic mechanism (n~=~17)\\
\textbullet~No substantive medical or clinically relevant biomedical application (n~=~23)\\
\textbullet~Ineligible publication type (n~=~45)\\
\textbullet~Insufficient methodological or evaluation detail (n~=~16)\\
\textbullet~Duplicate or substantially overlapping report (n~=~7)};
\node[box, line width=1.2pt, text width=4.0cm, align=center, anchor=north] (inc) at ([yshift=-1.4cm]e1.south) {\textbf{Unique studies provisionally included in systematic evidence mapping (n~=~557)}};
\draw[ln, dashed] (id1.east |- epmc.west) -- (epmc.west);
\draw[arr] (id1) -- (s1);
\draw[arr] (s1) -- (dup);
\draw[arr] (s1) -- (s2);
\draw[arr] (s2) -- (ex1);
\draw[arr] (s2) -- (e0);
\draw[arr] (oth1) -- (oth2);
\draw[ln] (e0.east) -- (join);
\draw[ln] (oth2.west) -- (join);
\draw[arr] (join) -- (e1.north);
\draw[arr] (e1.east) -- (e1.east -| ex2.west);
\draw[arr] (e1) -- (inc);
\coordinate (px0) at (0.1,0);
\coordinate (px1) at (0.75,0);
\node[phase, fit={(px0 |- id1.north) (px1 |- id1.south)}] (p1) {};
\node[phasetext] at (p1.center) {Identification};
\node[phase, fit={(px0 |- s1.north) (px1 |- s2.south)}] (p2) {};
\node[phasetext] at (p2.center) {Screening};
\node[phase, fit={(px0 |- e0.north) (px1 |- e1.south)}] (p3) {};
\node[phasetext] at (p3.center) {Eligibility};
\node[phase, fit={($(px0 |- inc.north)+(0,0.15)$) (px1 |- inc.south) (px1 |- ex2.south)}] (p4) {};
\node[phasetext] at (p4.center) {Included};
\end{tikzpicture}
    \caption{Study identification and selection process. Searches of PubMed, IEEE Xplore, Europe PMC, arXiv, and medRxiv identified 2,192 records, and backward and forward citation searching yielded five additional reports. Of the 1,747 records available from the four exportable sources, 98 duplicate or substantially overlapping reports were removed, leaving 1,649 records for title-and-abstract screening. A total of 665 reports underwent eligibility assessment, of which 108 were excluded for predefined reasons, resulting in a provisional evidence-mapping set of 557 unique studies. Europe PMC records were included in the identification-stage total but could not be independently exported, deduplicated, or screened at the individual-record level.}
    \label{fig:study_selection}
\end{figure}


%
\subsection{Study Selection}

Records retrieved from PubMed, IEEE Xplore, arXiv, and medRxiv were combined into a single screening dataset. Duplicate and substantially overlapping reports were identified using persistent identifiers, including DOI, PMID, and arXiv identifiers, together with normalized title matching. When multiple reports described the same underlying study, the most complete or formally published version was retained.

Europe PMC records could not be exported at the individual-record level and were therefore retained only in the identification-stage count, without record-level deduplication or independent screening. Records available from the four exportable sources underwent title-and-abstract screening, and potentially eligible reports proceeded to provisional eligibility assessment. Reports identified through backward reference screening and forward citation tracking were assessed using the same criteria. The PRISMA-ScR flow diagram in Figure~\ref{fig:study_selection} summarizes the overall study identification and selection process.

Full-text eligibility was determined from the implemented system architecture, workflow, and evaluation rather than from the use of terms such as ``agent'', ``AI agent'', or ``Agentic AI''. Assessment focused on goal-directed multistep execution, task planning or decomposition, external tool or code use, interaction with external resources, state or memory maintenance, feedback-based refinement, and multi-agent collaboration. Each report excluded during the provisional eligibility assessment was assigned one primary reason for exclusion: absence of an explicit agentic mechanism, lack of a substantive medical or clinically relevant biomedical application, ineligible publication type, insufficient methodological or evaluation detail, or duplication of a more complete or substantially overlapping report.

\subsection{Data Synthesis and Taxonomy Construction}

Given the substantial variation in terminology, task design, system implementation, data sources, and evaluation methods, we synthesized the evidence using narrative synthesis and systematic categorical mapping rather than quantitative meta-analysis. For each eligible study, we extracted the publication year and type; medical application domain; primary task; input modality; data source and dataset; intended user or clinical setting; underlying foundation model or controller; agent architecture; number and roles of agents; planning or task-decomposition mechanism; use of external tools, code, databases, or models; retrieval and external-knowledge integration; memory or state maintenance; feedback, reflection, or error-correction mechanisms; multimodal processing; human oversight; evaluation dataset; reported task and process metrics; expert involvement; validation setting; and publication-version relationship. We also recorded uncertainty handling, evidence traceability, safety controls, failure reporting, workflow integration, and clinical deployment where this information was available.

Studies were mapped across five analytical dimensions: agentic capability, technical architecture, medical application, evaluation strategy, and level of clinical validation. Agentic capabilities included goal-directed execution, planning, tool use, environmental interaction, memory, feedback-based refinement, and multi-agent collaboration. Technical architectures were classified as single-agent tool-use systems, multi-agent collaboration systems, external knowledge-augmented systems, or multimodal medical agents. Code execution was recorded as a separate implementation feature because it could occur across multiple architectural categories. Studies could therefore be assigned more than one implementation characteristic where appropriate.

Medical applications were classified into medical image interpretation and report generation, multimodal image analysis and downstream task coordination, diagnostic assistance and clinical reasoning, electronic health record analysis, drug safety and prescription review, clinical trial and biomedical research applications, and other clinically relevant workflows. Evaluation approaches were grouped into task performance, process reliability, evidence grounding and traceability, safety and uncertainty, usability and workflow impact, and external or clinical validation. Task-performance measures included accuracy, area under the receiver operating characteristic curve, sensitivity, specificity, precision, recall, F1-score, Dice coefficient, intersection over union, Hausdorff distance, BLEU, ROUGE, and BERTScore, where applicable. Process-oriented measures included task-completion rate, tool-use success rate, tool-selection accuracy, code-execution success, retrieval relevance, evidence consistency, multistep reasoning quality, error recovery, abstention or escalation behavior, and human revision burden.

The taxonomy and evidence map were refined iteratively through comparison of study-level characteristics across the extracted dimensions. The taxonomy was used to identify recurrent architectural patterns and application areas, whereas the evidence map summarized the relationship between agentic complexity and the maturity of clinical validation.

The extracted items, their analytical purpose, and their relationship to the mapping dimensions are summarized in Table~\ref{tab:extraction}.

\begin{table}[!htbp]
\centering
\caption{Data extraction framework used for evidence mapping. Items were extracted where reported in the original studies and were used to classify architectures, applications, evaluation strategies, and validation maturity. Missing items were recorded as not reported rather than inferred.}
\label{tab:extraction}
\footnotesize
\renewcommand{\arraystretch}{1.15}
\begin{tabularx}{\textwidth}{
>{\raggedright\arraybackslash}p{3.0cm}
>{\raggedright\arraybackslash}X
>{\raggedright\arraybackslash}p{4.4cm}}
\toprule
Domain & Extracted items & Analytical purpose \\
\midrule
Study characteristics & Publication year and type; publication-version relationship; data source and dataset & Study identification, consolidation of overlapping reports, and contextualization \\
Clinical context & Medical application domain; primary task; intended user or clinical setting; input modality & Application mapping and characterization of intended use \\
System design & Foundation model or controller; agent architecture; number and roles of agents; planning or task-decomposition mechanism & Classification of technical architecture \\
Agentic mechanisms & Use of external tools, code, databases, or models; retrieval and external-knowledge integration; memory or state maintenance; feedback, reflection, or error correction; multimodal processing & Capability mapping and derivation of the descriptive agentic complexity score \\
Oversight and safety & Human oversight; uncertainty handling; evidence traceability; safety controls; failure reporting & Appraisal of auditability, safety, and responsibility boundaries \\
Evaluation & Evaluation dataset; task and process metrics; expert involvement & Mapping of evaluation strategies across the six evaluation dimensions \\
Validation and deployment & Validation setting; workflow integration; clinical deployment & Classification of clinical validation maturity \\
\bottomrule
\end{tabularx}
\end{table}

\begin{table}[!t]
\centering
\caption{Overview of the scoping review methodology and systematic evidence-mapping process. The table links each review stage to its principal analytical focus and its role in defining the review scope, identifying eligible studies, extracting study characteristics, organizing the evidence, and assessing clinical validation maturity. Study selection was based on goal-directed multistep execution and explicit agentic mechanisms rather than on agent-related terminology alone. Detailed search, eligibility, selection, and synthesis procedures are described in the corresponding Methods subsections.}
\label{tab:review_design}
\small
\renewcommand{\arraystretch}{1.18}
\begin{tabularx}{\textwidth}{
>{\raggedright\arraybackslash}p{3.0cm}
>{\raggedright\arraybackslash}X
>{\raggedright\arraybackslash}X}
\toprule
Review stage & Main focus & Role in the review \\
\midrule
Review design &
Scoping review with a systematic mapping component
\cite{Arksey2005Scoping,Tricco2018PRISMAScR,Petersen2008Mapping} &
Research boundary, categorical mapping, and evidence organization \\

Search strategy &
Medical AI agents; LLM-based agents; Jan 2022--Jul 2026 &
Initial literature pool \\

Study selection &
Goal orientation; tool use; multi-step workflow &
Eligible agentic AI studies \\

Data extraction &
Architecture; application; evaluation method &
Structured evidence summary \\

Evidence mapping &
Agentic complexity; five-stage validation setting &
Capability--validation gap and clinical maturity \\
\bottomrule
\end{tabularx}
\end{table}

\subsection{Evidence Appraisal and Clinical Validation Maturity}

Given the substantial heterogeneity in study design, task type, data source, and evaluation setting, no single conventional risk-of-bias instrument was applicable across all included studies. We therefore appraised the evidence descriptively across predefined domains rather than assigning a single composite quality rating. These domains included clarity of the clinical task and intended use; provenance and representativeness of the evaluation data; involvement of clinicians or other domain experts; reporting of failure cases and error patterns; uncertainty or abstention mechanisms; evidence traceability; human oversight; workflow integration; and external or multi-institutional validation. For medical imaging studies, additional consideration was given to lesion localization, segmentation evidence, image--text consistency, and whether generated interpretations were supported by identifiable visual findings.

Clinical validation maturity was classified into five ordered levels. \textit{Static benchmark} included evaluations on static public datasets, medical question-answering datasets, or other offline tasks. \textit{Agent benchmark} included evaluations in interactive, tool-using, or simulated workflow environments developed specifically for medical agents. \textit{Retrospective real data} referred to evaluations using previously collected clinical records, electronic health records, medical images, or institutional datasets. \textit{External evaluation} included external-dataset testing, independent expert assessment, reader studies, and cross-institutional evaluation. \textit{Prospective workflow} encompassed prospective studies, silent deployment, real-world clinical workflow testing, and clinical trials conducted in operational healthcare settings.

When a study met the criteria for more than one adjacent level, classification was based on the principal validation design and the overall strength of the reported evidence. Higher levels indicated greater proximity to routine clinical practice but did not, in themselves, establish clinical effectiveness, safety, regulatory readiness, or generalizability. Agentic complexity was summarized using a descriptive score ranging from 0 to 8 according to the reported presence of planning, tool use, retrieval, memory, reflection, multi-agent collaboration, multimodal processing, and workflow integration. The score was used solely for comparative visualization and evidence mapping and was not treated as a validated measure of agentic capability, system quality, or clinical readiness. The principal stages of the review methodology, their analytical focus, and their roles in evidence synthesis are summarized in Table~\ref{tab:review_design}.

\section{Results}

\subsection{Search and Selection Results}


%
Searches across the five electronic sources identified 2,192 records, and backward and forward citation searching yielded five additional reports. Of the 1,747 records retrieved from the four sources that permitted record-level export, 98 duplicate or substantially overlapping reports were removed, leaving 1,649 records for title-and-abstract screening. After exclusion of 989 records, 660 reports proceeded to eligibility assessment. The 445 Europe PMC records contributed to the identification-stage total but could not be exported, deduplicated, or screened at the individual-record level. Together with the five reports identified through citation searching, 665 reports underwent eligibility assessment.

During the provisional eligibility assessment, 108 reports were excluded: 17 lacked an explicit agentic mechanism, 23 lacked a substantive medical or clinically relevant biomedical application, 45 were ineligible publication types, 16 provided insufficient methodological or evaluation detail, and seven were duplicate or substantially overlapping reports. The remaining 557 unique studies formed the provisional evidence-mapping set.

\subsection{Current Research Landscape}

Medical Agentic AI has emerged at the intersection of large language models, multimodal foundation models, and agent-oriented system design. Earlier medical AI systems, including rule-based and reinforcement-learning approaches, supported automation and sequential decision-making in narrowly defined settings. Reinforcement learning has been investigated in healthcare for sequential decision-making, although its predefined action spaces, optimization objectives, and safety constraints differ from those of contemporary LLM-based agentic systems \cite{Gottesman2019RLHealthcare}. Advances in language understanding, task planning, multimodal reasoning, and tool use have subsequently enabled systems to move beyond isolated prediction toward interactive and multistep medical workflows.

Earlier autonomous medical AI systems also demonstrated that clinically relevant tasks could be automated within tightly specified boundaries. For example, an autonomous diagnostic system was evaluated for diabetic retinopathy detection in primary care \cite{Abramoff2018AutonomousDR}. Such systems were designed around a predefined diagnostic objective and execution pathway and should therefore be distinguished from contemporary agentic systems that incorporate open-ended planning, dynamic tool use, external resource interaction, or iterative task execution.

The longer history of medical AI clarifies which elements of contemporary agentic systems are genuinely new. Early expert systems, such as MYCIN for antimicrobial therapy consultation and INTERNIST-I for diagnosis in internal medicine, already combined goal-directed inference with explicit knowledge bases and explanation facilities \cite{Shortliffe1975MYCIN,Miller1982INTERNIST}. Their reasoning was transparent but brittle because knowledge had to be encoded and maintained manually, and their descendants persist mainly as rule-based clinical decision support embedded in health information systems \cite{Sutton2020CDSS}. Classical machine learning and, subsequently, deep learning shifted the emphasis from encoded rules to learned representations \cite{LeCun2015DeepLearning,Rajkomar2019MLMedicine}. Deep learning achieved specialist-level performance on well-defined perceptual tasks, including diabetic retinopathy screening, skin lesion classification, retinal disease referral, lung cancer screening, breast cancer screening, and chest radiograph triage \cite{Gulshan2016DiabeticRetinopathy,Esteva2017SkinCancer,DeFauw2018RetinalDisease,Ardila2019LungCancer,McKinney2020BreastCancer,Annarumma2019ChestRadiographs}. Parallel work on electronic health records produced predictive and representation models for longitudinal patient data \cite{Choi2016RETAIN,Rajkomar2018EHRDeepLearning,Li2020BEHRT,Rasmy2021MedBERT}, whereas reinforcement learning was explored for sequential treatment decisions such as sepsis management \cite{Komorowski2018AIClinician,Yu2023RLHealthcareSurvey}. These systems were nevertheless typically confined to a single modality, a fixed output space, and a predefined position in the clinical workflow.

The foundation-model era relaxed these constraints. Domain-specific language models, including BioBERT, ClinicalBERT, PubMedBERT, and GatorTron, demonstrated the value of large-scale biomedical and clinical pretraining \cite{Lee2020BioBERT,Alsentzer2019ClinicalBERT,Gu2021PubMedBERT,Yang2022EHRLanguageModel}, and the broader concept of foundation models described large pretrained models that can be adapted to many downstream tasks \cite{Bommasani2021FoundationModels,Moor2023GMAI}. Instruction-tuned and preference-aligned LLMs subsequently reached or exceeded passing thresholds on medical licensing-style examinations and approached expert-level performance on several medical question-answering benchmarks \cite{Ouyang2022InstructGPT,OpenAI2023GPT4,Kung2023ChatGPTUSMLE,Nori2023GPT4Medical,Singhal2023ClinicalKnowledge,Singhal2025ExpertMedicalQA}. Agentic AI builds on these capabilities by embedding foundation models within control loops that plan, call tools, retrieve evidence, and revise actions. Representative milestones along this trajectory are summarized in Table~\ref{tab:milestones}. The transition is cumulative rather than substitutive: rule-based checks, task-specific deep learning models, and foundation models frequently reappear as components within agentic workflows.

\begin{table}[!htbp]
\centering
\caption{Representative milestones in the development of medical Agentic AI. Periods are approximate and reflect publication dates of representative works rather than the first appearance of each idea. The milestones illustrate the cumulative integration of knowledge representation, learned perception, foundation models, tool use, and interactive evaluation; they do not indicate clinical readiness.}
\label{tab:milestones}
\footnotesize
\renewcommand{\arraystretch}{1.15}
\begin{tabularx}{\textwidth}{
>{\raggedright\arraybackslash}p{1.6cm}
>{\raggedright\arraybackslash}p{4.1cm}
>{\raggedright\arraybackslash}X}
\toprule
Period & Milestone & Relevance to medical Agentic AI \\
\midrule
1975--1982 & Rule-based consultation systems (MYCIN, INTERNIST-I) \cite{Shortliffe1975MYCIN,Miller1982INTERNIST} & Goal-directed inference with explicit knowledge and explanation, but manual knowledge engineering \\
2016--2020 & Deep learning for diagnostic perception \cite{Gulshan2016DiabeticRetinopathy,Esteva2017SkinCancer,DeFauw2018RetinalDisease,Ardila2019LungCancer,McKinney2020BreastCancer} & Specialist-level single-task models that now serve as callable perception tools \\
2018 & Autonomous diabetic retinopathy detection; reinforcement learning for sepsis treatment \cite{Abramoff2018AutonomousDR,Komorowski2018AIClinician} & Autonomy within a fixed task and prospective validation; sequential decision-making with predefined action spaces \\
2019--2022 & Biomedical and clinical language models \cite{Lee2020BioBERT,Alsentzer2019ClinicalBERT,Gu2021PubMedBERT,Yang2022EHRLanguageModel} & Domain-specific language representations for clinical text \\
2021--2023 & Foundation models and generalist medical AI \cite{Bommasani2021FoundationModels,Moor2023GMAI,Tu2024GeneralistBiomedicalAI} & Adaptable multitask and multimodal models as controllers and components \\
2022--2023 & Reasoning and acting with language models \cite{Wei2022ChainOfThought,Yao2023ReAct,Schick2023Toolformer,Shinn2023Reflexion} & Interleaving of reasoning, tool calls, and feedback-based revision \\
2023 & LLM performance on medical examinations and clinical knowledge \cite{Kung2023ChatGPTUSMLE,Nori2023GPT4Medical,Singhal2023ClinicalKnowledge} & Evidence that general and medical LLMs encode substantial clinical knowledge \\
2023--2024 & General multi-agent and tool-orchestration frameworks \cite{Li2023CAMEL,Shen2023HuggingGPT,Wu2024AutoGen,Hong2024MetaGPT} & Reusable patterns for role specialization, communication, and model orchestration \\
2023--2024 & Tool-augmented scientific agents \cite{Boiko2023AutonomousChemicalResearch,Bran2024ChemCrow,Jin2024GeneGPT} & Autonomous planning and execution with domain tools and laboratory automation \\
2024 & First wave of medical agents \cite{Tang2024MedAgents,Kim2024MDAgents,Shi2024EHRAgent,Li2024MMedAgent,Li2024AgentHospital} & Multi-agent clinical reasoning, code-based EHR analysis, multimodal tool use, and simulated hospitals \\
2024--2026 & Interactive medical agent benchmarks \cite{Li2024MediQ,Jiang2025MedAgentBench,Zhu2025MedAgentBoard,Fan2025AIHospital,Schmidgall2026AgentClinic,Johri2025CRAFTMD} & Evaluation of information gathering, tool use, and collaboration beyond static questions \\
2025 & Peer-reviewed agent studies with expert or experimental validation \cite{Tu2025ConversationalDiagnosticAI,Ferber2025OncologyAgent,Swanson2025VirtualLab,Qu2025CRISPRGPT} & Diagnostic dialogue, oncology decision support, and wet-laboratory validation of agent-designed outputs \\
2025--2026 & Reasoning models and cost-aware, rubric-based evaluation \cite{Guo2025DeepSeekR1,Nori2026SequentialDiagnosis,Arora2025HealthBench} & Reinforcement-learned reasoning, sequential diagnosis with testing costs, and physician-written evaluation rubrics \\
\bottomrule
\end{tabularx}
\end{table}

The evidence identified in this review remains concentrated in proof-of-concept systems, benchmark studies, simulated environments, retrospective datasets, and small-scale expert evaluations. Applications include medical question answering, report generation, image interpretation, lesion detection and segmentation, medical visual question answering, multimodal tool use, clinical decision support, and clinical trial prediction. These studies provide early evidence of technical feasibility, particularly for coordinating specialized models and information sources, but offer limited evidence of prospective clinical effectiveness, workflow benefit, or generalizability. 
Medical Agentic AI is therefore more appropriately regarded as an emerging systems paradigm than as a clinically mature technology.

\begin{figure}[!t]
    \centering
    \includegraphics[width=\textwidth]{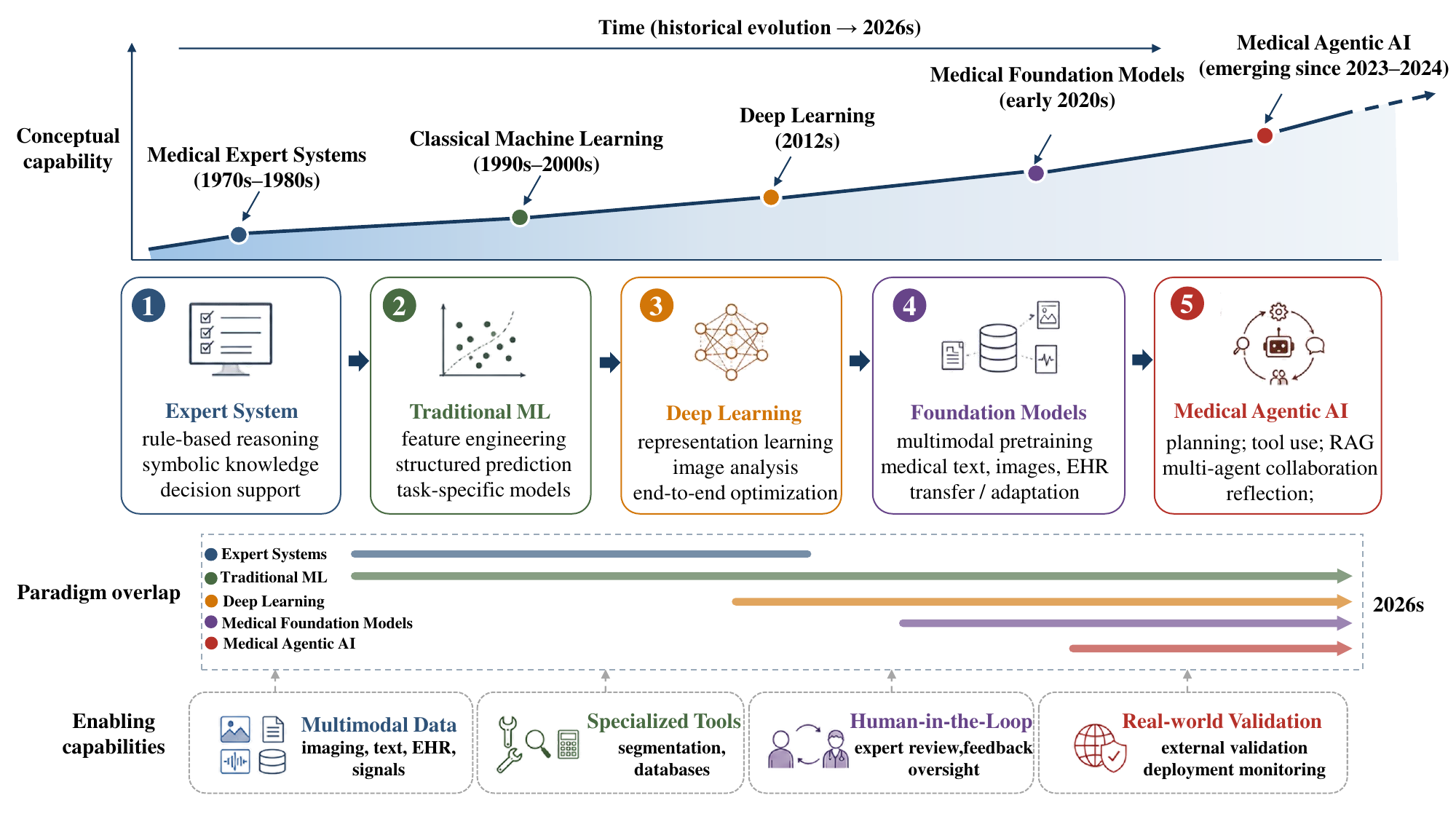}
\caption{Historical evolution of medical AI toward Agentic AI.
Schematic overview of the progression from expert systems and conventional machine learning to deep learning, medical foundation models, and emerging medical Agentic AI. The upper trajectory and horizontal bars indicate approximate periods of emergence, continued use, and overlap among paradigms, rather than quantitative gains in model performance or clinical maturity. The lower panel highlights multimodal data, specialized tools, human oversight, and real-world validation as cross-cutting enabling conditions for clinical translation. EHR, electronic health record; RAG, retrieval-augmented generation.
}
    \label{fig:medical_agentic_ai_framework}
\end{figure}

Previous reviews have reached broadly consistent conclusions while adopting different definitions and scopes. Collaco et al. applied a relatively restrictive definition of Agentic AI and included seven healthcare studies across emergency medicine, oncology, radiology, and rehabilitation, most of which were exploratory, experimental, or computational and lacked substantial real-world validation \cite{Collaco2026AgenticHealthcare}. Zhao et al. adopted a broader AI-agent perspective and reviewed applications in diagnostic assistance, clinical decision support, medical report generation, patient interaction, hospital management, and medical education \cite{Zhao2026AIAgentHealthcare}. Collectively, these reviews indicate an expanding application landscape but continued inconsistency in conceptual definitions, system architectures, evaluation standards, and pathways to clinical translation. Figure~\ref{fig:medical_agentic_ai_framework} situates this emerging paradigm within the broader evolution of medical AI and highlights the capabilities and enabling conditions associated with clinical translation.

\subsection{Definitions and Core Capabilities of Medical Agentic AI}

The terms AI agent and Agentic AI are used inconsistently across the medical literature. An AI agent generally denotes an autonomous computational entity capable of perceiving its environment, interpreting a task, selecting actions, and pursuing a defined objective \cite{Wooldridge1995Agents}. Agentic AI, by contrast, refers to the system-level orchestration of one or more agents within a goal-directed workflow that may involve multistep reasoning, memory, tool use, interaction with external resources, and adaptive revision. Under this distinction, an AI agent may function as an individual component, whereas an Agentic AI system may comprise either a single autonomous agent or multiple collaborating agents.

Drawing on surveys of LLM-based autonomous agents and emerging healthcare-specific taxonomies, we characterize medical Agentic AI through eight interrelated capabilities: goal orientation, reasoning and planning, memory, tool use, retrieval augmentation, feedback-based refinement, multi-agent collaboration, and simulation \cite{Wang2024LLMAgentsSurvey,Vatsal2026AgenticTaxonomy}. These capabilities need not be implemented as separate modules in every system. Rather, they describe the mechanisms through which an agentic system maintains task objectives, coordinates specialized resources, responds to intermediate results, and executes complex medical workflows. Figure~\ref{fig:medical_agentic_ai_overview} summarizes the relationships among multimodal clinical inputs, these core capabilities, and representative application domains.

Goal orientation defines the clinical or operational objective around which the workflow is organized, such as medical image interpretation, radiology report generation, electronic health record analysis, prescription review, or clinical decision support. Reasoning and planning translate this objective into a sequence of executable actions. In imaging-assisted diagnosis, for example, a system may identify the clinical question, invoke detection or segmentation tools, and integrate imaging findings with patient history, laboratory results, and previous reports. In electronic health record analysis, planning may involve retrieving relevant variables, performing temporal or tabular calculations, and comparing alternative diagnostic hypotheses. General approaches such as ReAct and Tree of Thoughts support reasoning through interaction with external actions or exploration of alternative reasoning paths, although their effectiveness and safety in medicine require task-specific validation \cite{Yao2023ReAct,Yao2023TreeOfThoughts}.

\begin{figure}[!t]
    \centering
\definecolor{fwblue}{RGB}{58,124,190}
\definecolor{fwgreen}{RGB}{96,160,70}
\definecolor{fwpurple}{RGB}{122,78,160}
\definecolor{fwgray}{RGB}{95,95,95}
\begin{tikzpicture}[
  font=\sffamily,
  every node/.append style={execute at begin node={\hyphenpenalty=10000\exhyphenpenalty=10000}},
  panel/.style={draw=#1, line width=0.9pt, rounded corners=5pt, fill=#1!5},
  ptitle/.style={font=\sffamily\small\bfseries, text=#1, align=center},
  inbox/.style={draw=fwblue!70, fill=white, rounded corners=2pt, minimum width=3.2cm, minimum height=1.15cm, text width=3.0cm, align=center, font=\sffamily\fontsize{7.5}{9}\selectfont, inner sep=2pt},
  capbox/.style={draw=fwgreen!80!black, fill=white, rounded corners=2pt, minimum width=3.0cm, minimum height=1.6cm, text width=2.85cm, align=center, font=\sffamily\fontsize{7.5}{9}\selectfont, inner sep=2pt},
  appbox/.style={draw=fwpurple!80, fill=white, rounded corners=2pt, minimum width=3.0cm, minimum height=1.1cm, text width=2.85cm, align=center, font=\sffamily\fontsize{7.5}{9}\selectfont\bfseries, text=fwpurple!65!black, inner sep=2pt},
  ctrl/.style={draw=fwgreen!80!black, fill=fwgreen!18, rounded corners=2pt, minimum width=12.7cm, minimum height=0.85cm, text width=12.4cm, align=center, font=\sffamily\fontsize{7.5}{9}\selectfont, inner sep=2pt},
  big/.style={-{Stealth[length=3mm,width=2.8mm]}, line width=2.2pt, draw=#1}
]
\draw[panel=fwblue] (0,0) rectangle (3.6,-9.35);
\node[ptitle=fwblue!70!black] at (1.8,-0.5) {Multimodal\\clinical inputs};
\draw[panel=fwgreen] (4.1,0) rectangle (17.4,-5.3);
\node[ptitle=fwgreen!45!black] at (10.75,-0.32) {Agent capabilities};
\node[ctrl] at (10.75,-1.07) {\textbf{Foundation-model controller:} task interpretation \textbullet{} planning \textbullet{} orchestration of tools, knowledge, and agents \textbullet{} integration of intermediate outputs};
\draw[panel=fwpurple] (4.1,-5.9) rectangle (17.4,-9.35);
\node[ptitle=fwpurple!65!black] at (10.75,-6.24) {Application ecosystem};
\node[inbox] at (1.8,-1.75) {{\bfseries\color{fwblue!65!black}Medical images}\\radiology, pathology};
\node[inbox] at (1.8,-3.05) {{\bfseries\color{fwblue!65!black}Clinical text}\\notes and reports};
\node[inbox] at (1.8,-4.35) {{\bfseries\color{fwblue!65!black}EHR and structured data}\\labs and medications};
\node[inbox] at (1.8,-5.65) {{\bfseries\color{fwblue!65!black}Physiological signals}\\ECG and vital signs};
\node[inbox] at (1.8,-6.95) {{\bfseries\color{fwblue!65!black}Genomic and molecular data}};
\node[inbox] at (1.8,-8.25) {{\bfseries\color{fwblue!65!black}Patient-reported information}};
\node[capbox] at (5.9,-2.5) {{\bfseries\color{fwgreen!45!black}Goal orientation}\\[1pt]Clinical objective and intended use};
\node[capbox] at (9.133,-2.5) {{\bfseries\color{fwgreen!45!black}Reasoning and planning}\\[1pt]Task decomposition and test selection};
\node[capbox] at (12.367,-2.5) {{\bfseries\color{fwgreen!45!black}Memory}\\[1pt]Case and patient context over time};
\node[capbox] at (15.6,-2.5) {{\bfseries\color{fwgreen!45!black}Tool use}\\[1pt]Calculators, imaging models, and code};
\node[capbox] at (5.9,-4.3) {{\bfseries\color{fwgreen!45!black}Retrieval augmentation}\\[1pt]Guidelines, literature, and drug labels};
\node[capbox] at (9.133,-4.3) {{\bfseries\color{fwgreen!45!black}Feedback-based refinement}\\[1pt]Error correction, abstention, and escalation};
\node[capbox] at (12.367,-4.3) {{\bfseries\color{fwgreen!45!black}Multi-agent collaboration}\\[1pt]Role specialization and consensus};
\node[capbox] at (15.6,-4.3) {{\bfseries\color{fwgreen!45!black}Simulation}\\[1pt]Simulated patients and digital twins};
\node[appbox] at (5.9,-7.2) {Image interpretation and report generation};
\node[appbox] at (9.133,-7.2) {Multimodal image analysis and task coordination};
\node[appbox] at (12.367,-7.2) {EHR analysis and clinical documentation};
\node[appbox] at (15.6,-7.2) {Diagnostic assistance and clinical reasoning};
\node[appbox] at (5.9,-8.55) {Treatment planning and disease management};
\node[appbox] at (9.133,-8.55) {Drug safety, trials, and biomedical research};
\node[appbox] at (12.367,-8.55) {Patient support and telehealth};
\node[appbox] at (15.6,-8.55) {Public health and medical education};
\draw[fill=fwgray!7, draw=fwgray, line width=0.8pt, rounded corners=5pt] (0,-9.85) rectangle (17.4,-10.75);
\node[font=\sffamily\fontsize{7.5}{9}\selectfont, text width=16.8cm, align=center] at (8.7,-10.3) {\textbf{Human oversight and safety boundary:} clinician review \textbullet{} action permissions \textbullet{} evidence traceability \textbullet{} uncertainty-aware abstention and escalation \textbullet{} audit logging and monitoring};
\draw[big=fwblue] (3.62,-3.4) -- (4.08,-3.4);
\draw[big=fwgreen!80!black] (10.75,-5.32) -- (10.75,-5.88);
\draw[{Stealth[length=2mm]}-{Stealth[length=2mm]}, line width=0.9pt, draw=fwgray] (1.8,-9.37) -- (1.8,-9.83);
\draw[{Stealth[length=2mm]}-{Stealth[length=2mm]}, line width=0.9pt, draw=fwgray] (10.75,-9.37) -- (10.75,-9.83);
\end{tikzpicture}
\caption{Conceptual framework of medical Agentic AI. Multimodal clinical inputs, including medical images, clinical text and reports, structured electronic health record data, physiological signals, genomic and molecular data, and patient-reported information, provide the information environment for agentic workflows. A foundation-model controller interprets the task and orchestrates eight core capabilities: goal orientation, reasoning and planning, memory, tool use, retrieval augmentation, feedback-based refinement, multi-agent collaboration, and simulation. These capabilities support applications in image interpretation and report generation, multimodal image analysis, electronic health record analysis and documentation, diagnostic assistance and clinical reasoning, treatment planning, drug safety and biomedical research, patient support and telehealth, and public health and medical education. Human oversight and safety mechanisms bound the workflow. The pathways are conceptual and do not imply that every system implements all capabilities or follows a fixed execution sequence.}
    \label{fig:medical_agentic_ai_overview}
\end{figure}

Memory maintains relevant context, intermediate outputs, previous tool calls, and human feedback across multistep or longitudinal tasks. Its role extends beyond preserving dialogue history and is particularly relevant to follow-up care, chronic disease management, imaging review, and multi-turn clinical reasoning. Tool use allows the agent to delegate specialized operations to external resources, including clinical databases, guideline repositories, drug databases, code interpreters, electronic health record interfaces, image-analysis models, and statistical tools. EHRAgent uses code execution to support tabular reasoning over electronic health records, whereas MMedAgent integrates multiple medical tools as callable components within a multimodal workflow \cite{Shi2024EHRAgent,Li2024MMedAgent}. The reliability of the resulting workflow therefore depends on appropriate tool selection, accurate interpretation of tool outputs, and the performance of the invoked components.

Retrieval augmentation provides access to information beyond the model's parametric knowledge, including clinical guidelines, drug information, medical literature, and previous cases. General retrieval-augmented generation methods support knowledge-intensive question answering, while MALADE illustrates how external evidence can be incorporated into a medical agentic workflow for pharmacovigilance \cite{Lewis2020RAG,Izacard2021PassageRetrieval,Choi2024MALADE}. The value of retrieval depends on source quality, relevance, recency, and correct interpretation; retrieval alone does not ensure factual accuracy or clinical safety \cite{Chen2025EvaluatingLLMAgentsHealthcare}. Feedback-based refinement complements retrieval and tool use by allowing the system to revise subsequent actions in response to intermediate outputs, execution failures, or human input. Methods such as Reflexion demonstrate how feedback can modify later reasoning and behavior \cite{Shinn2023Reflexion}. In medical settings, such mechanisms should support uncertainty recognition, error correction, abstention, and escalation to clinician review.

Multi-agent collaboration distributes specialized functions across agents with different roles or information sources. Medical systems may assign separate agents to image analysis, clinical reasoning, drug safety, report generation, or result verification. MedAgents and MDAgents use role-specialized interaction to support multidisciplinary reasoning and adaptive collaboration across medical tasks \cite{Tang2024MedAgents,Kim2024MDAgents}. Multi-agent architectures, however, are not inherently more reliable than single-agent systems. Their performance depends on task complexity, role definition, communication protocols, evidence arbitration, disagreement resolution, and control of error propagation \cite{Zhu2025MedAgentBoard}. Together, these capabilities define the functional scope of medical Agentic AI but do not, by themselves, establish clinical effectiveness, autonomy, or deployment readiness.

Simulation, the eighth capability, is of growing importance. Agents may interact with simulated patients, virtual hospitals, or patient-specific computational models to rehearse information gathering, compare alternative actions, or accumulate experience before acting in real settings \cite{Li2024AgentHospital,Schmidgall2026AgentClinic,Almansoori2025MedAgentSim}. Digital twins extend this idea by maintaining a continuously updated computational representation of an individual patient or care process that can be queried for what-if analysis \cite{Katsoulakis2024DigitalTwins}. In current medical Agentic AI, however, simulation is used predominantly for evaluation and training rather than for patient-specific decision-making, and the fidelity of simulated environments limits the transferability of the resulting evidence.

These capabilities can be combined with different degrees of autonomy. At one end of the spectrum, an agent retrieves information or drafts outputs that are always reviewed by a clinician; at intermediate levels, it executes bounded multistep tasks, such as querying records or running predefined analyses, within explicit permissions; and at the other end, it would initiate clinically consequential actions without case-by-case approval. The systems discussed in this review are predominantly evaluated as assistive or supervised tools rather than as autonomous actors in clinical care. Autonomy should therefore be specified as a property of a particular task, setting, and permission structure rather than as an intrinsic attribute of a model \cite{Gao2024BiomedicalAgents,Mehandru2024AgentsInClinic}. Tables~\ref{tab:concepts} and~\ref{tab:capabilities} summarize the conceptual distinctions and core capabilities discussed in this section.

\begin{table}[!htbp]
\centering
\caption{Medical Agentic AI and adjacent concepts. The distinctions are functional rather than architectural: the same foundation model may serve as a standalone assistant, a callable tool, or the controller of an agentic workflow, depending on how it is embedded and what actions it is permitted to take.}
\label{tab:concepts}
\footnotesize
\renewcommand{\arraystretch}{1.15}
\begin{tabularx}{\textwidth}{
>{\raggedright\arraybackslash}p{2.8cm}
>{\raggedright\arraybackslash}X
>{\raggedright\arraybackslash}p{3.6cm}
>{\raggedright\arraybackslash}p{3.4cm}}
\toprule
Concept & Defining characteristics & Representative medical examples & Relationship to Agentic AI \\
\midrule
Task-specific medical AI model & Maps a predefined input to a fixed output; no planning or external action & Retinopathy screening, lesion segmentation \cite{Gulshan2016DiabeticRetinopathy,Isensee2021nnUNet} & Frequently invoked as a specialized tool \\
Rule-based decision support & Deterministic rules, alerts, and order sets triggered by predefined conditions & Expert systems and EHR alerts \cite{Shortliffe1975MYCIN,Sutton2020CDSS} & Supplies verifiable checks and action constraints \\
Medical foundation model & Large pretrained model adaptable to many tasks through prompting or fine-tuning & Med-PaLM, BiomedGPT, CONCH \cite{Singhal2023ClinicalKnowledge,Zhang2024BiomedGPT,Lu2024CONCH} & Serves as controller, reasoning engine, or perception module \\
LLM-based assistant & Conversational question answering or drafting without autonomous use of external resources & Responses to patient questions \cite{Ayers2023ChatbotResponses} & Becomes agentic when goals, tools, memory, or feedback loops are added \\
AI agent & Perceives an environment, selects actions, and pursues an objective & Single LLM agent querying an EHR \cite{Wooldridge1995Agents,Shi2024EHRAgent} & Building block of agentic systems \\
Multi-agent system & Multiple role-specialized agents that communicate and coordinate & Multidisciplinary consultation \cite{Tang2024MedAgents,Kim2024MDAgents} & Architectural subclass of Agentic AI \\
Agentic AI system & Goal-directed orchestration of agents, models, tools, knowledge, memory, and human oversight across multistep workflows & Tool-orchestrating imaging and oncology agents \cite{Fallahpour2025MedRAX,Ferber2025OncologyAgent} & Subject of this review \\
\bottomrule
\end{tabularx}
\end{table}

\begin{table}[!htbp]
\centering
\caption{Core capabilities of medical Agentic AI, representative implementation mechanisms, medical examples, and principal evaluation concerns. Capabilities are not necessarily implemented as separate modules, and the presence of a capability does not by itself establish clinical benefit.}
\label{tab:capabilities}
\footnotesize
\renewcommand{\arraystretch}{1.15}
\begin{tabularx}{\textwidth}{
>{\raggedright\arraybackslash}p{2.5cm}
>{\raggedright\arraybackslash}X
>{\raggedright\arraybackslash}X
>{\raggedright\arraybackslash}p{3.6cm}}
\toprule
Capability & Representative mechanisms & Medical examples & Principal evaluation concern \\
\midrule
Goal orientation & Task specification, system instructions, and decomposition of clinical objectives \cite{Wang2024LLMAgentsSurvey} & Prescription verification; radiograph interpretation \cite{Phan2024RxStrategist,Fallahpour2025MedRAX} & Clarity of intended use and scope boundaries \\
Reasoning and planning & Chain-of-thought, self-consistency, ReAct, and tree search \cite{Wei2022ChainOfThought,Wang2023SelfConsistency,Yao2023ReAct,Yao2023TreeOfThoughts} & Differential diagnosis and sequential test selection \cite{McDuff2025DifferentialDiagnosis,Nori2026SequentialDiagnosis} & Plan validity, information-gathering efficiency, and cost \\
Memory & Memory streams, hierarchical memory, and modular memory architectures \cite{Park2023GenerativeAgents,Packer2023MemGPT,Sumers2024CoALA} & Experience accumulation in simulated encounters; multidisciplinary case memory \cite{Almansoori2025MedAgentSim,Chen2026RareAgents} & Currency, provenance, privacy, and error persistence \\
Tool use & Learned API calling, large-scale tool learning, and code as action \cite{Schick2023Toolformer,Qin2024ToolLLM,Patil2024Gorilla,Wang2024CodeAct} & EHR code execution, clinical calculators, and genomic APIs \cite{Shi2024EHRAgent,Jin2025AgentMD,Jin2024GeneGPT} & Tool-selection accuracy, argument validity, and execution success \\
Retrieval augmentation & Retrieval-augmented generation and passage retrieval \cite{Lewis2020RAG,Izacard2021PassageRetrieval} & Clinical question answering and radiology retrieval \cite{Zakka2024Almanac,Xiong2024MedRAG,TayebiArasteh2025RadioRAG} & Source authority, temporal validity, and citation fidelity \\
Feedback-based refinement & Verbal reinforcement, self-refinement, and self-reflection \cite{Shinn2023Reflexion,Madaan2023SelfRefine} & Retrieval with self-reflection for biomedical reasoning \cite{Jeong2024SelfBioRAG} & Error detection, recovery, abstention, and escalation \\
Multi-agent collaboration & Role-playing, debate, and standardized operating procedures \cite{Li2023CAMEL,Du2024MultiagentDebate,Hong2024MetaGPT,Wu2024AutoGen} & Multidisciplinary consultation and EHR-based collaboration \cite{Tang2024MedAgents,Kim2024MDAgents,Wang2025ColaCare} & Independence of evidence, consensus validity, and cost \\
Simulation & Simulated patients, virtual hospitals, and digital twins \cite{Katsoulakis2024DigitalTwins} & Simulated clinical environments \cite{Li2024AgentHospital,Schmidgall2026AgentClinic,Johri2025CRAFTMD} & Simulation fidelity and transfer to real care \\
\bottomrule
\end{tabularx}
\end{table}

\subsection{Technical Architectures of Medical Agentic AI}

%
Medical Agentic AI is best understood as an orchestration paradigm rather than a single model class. Its defining feature is the organization of foundation models, specialized tools, external knowledge, memory, and human oversight within a goal-directed workflow. Unlike conventional end-to-end systems, agentic architectures decompose complex tasks into executable steps, select appropriate resources, maintain intermediate states, and revise subsequent actions in response to new evidence. Across the included literature, four recurrent architecture families can be identified: single-agent tool-use systems, multi-agent collaboration systems, knowledge-augmented agents, and multimodal medical agents. 
Code execution is treated as a cross-cutting implementation pattern because it may be incorporated into several of these architectures.

\subsubsection{Single-Agent Tool-Use Architecture}

Single-agent tool-use systems employ a large language model or multimodal foundation model as the central controller. The controller interprets the task, plans the execution sequence, selects external tools, and integrates their outputs into a final response. Callable resources may include code interpreters, clinical databases, guideline repositories, drug databases, electronic health record interfaces, image classification or segmentation models, report-generation modules, and statistical tools. EHRAgent, for example, uses code execution to support structured reasoning over electronic health records, whereas MMedAgent and MedRAX coordinate specialized medical tools within multimodal workflows \cite{Shi2024EHRAgent,Li2024MMedAgent,Fallahpour2025MedRAX}.

This architecture provides a relatively explicit control path because planning and tool selection remain concentrated in a single controller. It is therefore well suited to tasks with clearly defined objectives, such as structured EHR querying, image-based question answering, literature retrieval, and report assistance. EHR-oriented systems are frequently developed and retrospectively evaluated using resources such as MIMIC-IV, which supports clinical information extraction, temporal reasoning, and cohort analysis \cite{Johnson2023MIMICIV}. Performance on a public retrospective dataset, however, should not be equated with prospective operation within a hospital information system. The principal limitations of single-agent architectures arise from dependence on the controller's interpretation of the task, appropriate tool selection, and the reliability of each invoked component. Errors in planning, retrieval, code execution, or image analysis may propagate through the workflow and affect the final output.

\subsubsection{Multi-Agent Collaboration Architecture}

Multi-agent architectures distribute functions across agents with distinct roles, knowledge sources, or responsibilities. Their central mechanisms include role specialization, inter-agent communication, coordination, and consensus formation \cite{Guo2024MultiAgentsSurvey}. In medicine, such architectures may represent different clinical specialties or separate technical functions, including image interpretation, diagnostic reasoning, drug review, report generation, and result verification. MedAgents assigns medical expert roles to multiple agents and integrates their conclusions through iterative communication, whereas MDAgents adapts the collaboration strategy to task complexity by selecting between single-agent and multi-agent reasoning \cite{Tang2024MedAgents,Kim2024MDAgents}.

The potential advantage of this design lies in its ability to separate complementary forms of expertise and expose disagreement between agents. Its value, however, depends on whether role assignments correspond to meaningful task requirements and whether communication produces additional evidence rather than repeated model outputs. Available benchmark evidence does not show consistent superiority over strong single-model or specialized baselines \cite{Zhu2025MedAgentBoard}. Multi-agent systems also increase computational cost and introduce additional pathways for error propagation, unsupported consensus, and coordination failure. Reliable implementations therefore require explicit communication protocols, evidence arbitration, disagreement resolution, and mechanisms for escalating unresolved cases to human review.

\subsubsection{External Knowledge-Augmented Architecture}

Knowledge-augmented architectures connect an agent to information beyond the model's parametric knowledge. Relevant sources include clinical guidelines, medical literature, drug labels, knowledge graphs, institutional protocols, and previous cases. The agent retrieves evidence relevant to the task and incorporates it into subsequent reasoning or response generation. MALADE and Rx Strategist illustrate this pattern in pharmacovigilance and prescription-related reasoning through retrieval, structured external knowledge, and staged analysis \cite{Choi2024MALADE,Phan2024RxStrategist}.

External knowledge can improve currency and traceability, particularly where clinical recommendations or drug information change over time. It may also support image interpretation by linking visual findings to radiological terminology, disease knowledge, or reporting conventions. Nevertheless, retrieval should be regarded as an evidence-access mechanism rather than a guarantee of correctness. System performance depends on source authority, retrieval relevance, temporal validity, citation fidelity, and appropriate interpretation of the retrieved material. Irrelevant or outdated evidence may be propagated as confidently as valid information, making source-level provenance and task-specific evaluation essential.

\subsubsection{Multimodal Medical Agent Architecture}

Multimodal medical agents integrate images, text, structured clinical data, and specialized analytical tools within a common workflow. This architecture is particularly relevant to medicine because clinical interpretation often requires joint consideration of computed
tomography (CT), magnetic resonance imaging (MRI), positron emission tomography
(PET), radiographs, pathology findings, laboratory measurements, reports, and
longitudinal electronic health records. In contrast to task-specific imaging models, multimodal agents organize these information sources around a clinical question and may coordinate perception, localization, reasoning, and report generation.

Medical vision--language foundation models and tool-using multimodal agents occupy different positions within this architecture. CheXagent primarily provides chest radiograph understanding, visual question answering, abnormality recognition, and report-related capabilities and can therefore serve as a perception and semantic reasoning component within a broader workflow \cite{Chen2024CheXagent}. MMedAgent and MedRAX more explicitly implement agentic orchestration by selecting and integrating specialized imaging and text-processing tools according to task requirements \cite{Li2024MMedAgent,Fallahpour2025MedRAX}. Major limitations include cross-modal misalignment, inconsistency between generated text and image evidence, propagation of errors from detection or segmentation modules, and reduced generalizability across institutions with different equipment, acquisition protocols, data quality, and reporting practices. Evaluation should therefore include visual grounding, image--text consistency, uncertainty, and clinician-reviewable evidence rather than task performance alone.

\subsubsection{Foundation Model Backbones and Reasoning Engines}

The controller of most medical agents is a general-purpose or domain-adapted foundation model. Proprietary general-purpose LLMs, such as GPT-4 and Gemini-family models, are frequently used because of their instruction following, long-context processing, and native function-calling capabilities \cite{OpenAI2023GPT4,Saab2024MedGemini}. Med-Gemini illustrates how such models can be specialized for medicine and equipped with web search, thereby linking model-level capability with tool use \cite{Saab2024MedGemini}. Medical LLMs such as Med-PaLM and Med-PaLM~2 demonstrated the value of medical alignment and expert evaluation \cite{Singhal2023ClinicalKnowledge,Singhal2025ExpertMedicalQA}, whereas open models such as MEDITRON and PMC-LLaMA enable local deployment, inspection, and domain-specific fine-tuning, which are important when patient data cannot leave institutional infrastructure \cite{Chen2023Meditron,Wu2024PMCLLaMA}.

Multimodal foundation models extend this backbone to images. Biomedical vision--language assistants, including LLaVA-Med, Med-Flamingo, and BiomedGPT, support image-grounded dialogue and multitask biomedical analysis \cite{Li2023LLaVAMed,Moor2023MedFlamingo,Zhang2024BiomedGPT}, and MedVersa was trained on tens of millions of medical instances to achieve performance competitive with specialized models across multiple imaging tasks \cite{Zhou2024MedVersa}. RadFM, M3D, and models developed from the CT-RATE dataset address radiological and volumetric inputs \cite{Wu2025RadFM,Bai2024M3D,Hamamci2026CTRATE}, whereas pathology and ophthalmology foundation models, such as CONCH, UNI, PathChat, and RETFound, provide domain-specific representations and conversational interfaces \cite{Lu2024CONCH,Chen2024UNI,Lu2024PathChat,Zhou2023RETFound}. Contrastive biomedical encoders, such as BiomedCLIP, and prompt-driven parsing models, such as BiomedParse, are often better suited to serve as callable perception tools than as autonomous controllers \cite{Zhang2025BiomedCLIP,Zhao2025BiomedParse}.

A more recent development is the use of reasoning models trained with reinforcement learning to generate extended intermediate reasoning before answering \cite{Guo2025DeepSeekR1}. On a benchmark of difficult clinical questions, internalized reasoning models and externalized agent-based scaffolding each independently improved performance, suggesting that model-level reasoning and system-level orchestration are complementary rather than interchangeable \cite{Shao2026MedicalAgentsBench}. The choice of backbone nevertheless involves trade-offs among accuracy, cost, latency, privacy, controllability, version stability, and the feasibility of local validation \cite{Kapoor2025AgentsThatMatter}. Because proprietary models may change without notice, studies should report model versions, access dates, and decoding settings to support reproducibility.

\subsubsection{Tool Interfaces, Interoperability, and Execution Environments}

Tool use requires an interface through which the controller selects functions, supplies arguments, and receives outputs. Toolformer showed that language models can learn when to call external APIs, whereas ToolLLM and Gorilla scaled tool learning to large API collections and retrieval of API documentation \cite{Schick2023Toolformer,Qin2024ToolLLM,Patil2024Gorilla}. HuggingGPT used an LLM as a controller that dispatches subtasks to specialized models, a pattern closely related to medical tool orchestration \cite{Shen2023HuggingGPT}. In biomedicine, GeneGPT taught LLMs to use genomic web APIs, and AgentMD curated more than 2,000 executable clinical calculators from the literature and selected among them for patient-level risk prediction \cite{Jin2024GeneGPT,Jin2025AgentMD}.

Code generation offers a more expressive action space. Rather than selecting from a fixed set of functions, an agent may write and execute programs for data retrieval, statistical analysis, or modeling, as illustrated by general code-action agents and by EHRAgent \cite{Wang2024CodeAct,Shi2024EHRAgent}. An LLM-based code interpreter has also developed machine-learning models for real-world clinical datasets with performance comparable to that of published models \cite{TayebiArasteh2024AutoML}, and MedAgentGym provides executable sandboxed environments for training and evaluating agents on code-based biomedical reasoning tasks \cite{Xu2025MedAgentGym}. Executable actions, however, increase the need for sandboxing, input validation, data-access control, and independent verification of statistical validity.

Integration with clinical information systems is a prerequisite for operation in routine care. Fast Healthcare Interoperability Resources (FHIR) and SMART on FHIR provide standardized resources, application programming interfaces, and authorization mechanisms for EHR access \cite{Mandel2016SMARTonFHIR}, and MedAgentBench evaluates agents through FHIR-compliant interfaces in a virtual EHR environment \cite{Jiang2025MedAgentBench}. Emerging general-purpose protocols, such as the Model Context Protocol, aim to standardize connections between models and external tools or data sources, but they also introduce new attack surfaces across the life cycle of tool servers, including malicious tools and indirect prompt injection \cite{Hou2026MCP,Greshake2023IndirectInjection}. Interoperability should therefore be accompanied by least-privilege permissions, authenticated tool registries, and complete logging of tool calls \cite{Chan2024VisibilityAgents}.

\subsubsection{Memory and State Management}

Memory determines how an agent maintains task state across steps and sessions. Generative agents introduced memory streams combined with retrieval, reflection, and planning, and MemGPT proposed hierarchical memory management for contexts that exceed the model's context window \cite{Park2023GenerativeAgents,Packer2023MemGPT}; cognitive-architecture frameworks further distinguish working, episodic, semantic, and procedural memory \cite{Sumers2024CoALA}. Medical systems have used memory to accumulate experience across simulated encounters, retrieve knowledge from previous cases, and maintain multidisciplinary case information \cite{Li2024AgentHospital,Almansoori2025MedAgentSim,Chen2026RareAgents}. Clinical memory nevertheless raises specific concerns: retained information may become outdated, conflict with the current record, contain protected health information, or propagate earlier errors. Memory content should therefore be versioned, attributable to its source, and subject to the same access controls as the underlying clinical data.

\subsubsection{Integrated Agentic Workflows}

In practice, the architecture families described above are rarely mutually exclusive. Many systems combine a foundation model as the central reasoning or orchestration component with specialized medical tools, external knowledge, multimodal inputs, and human oversight. The foundation model interprets task requirements and integrates intermediate outputs, while specialized components perform narrower and more verifiable operations such as lesion detection, image segmentation, structured EHR analysis, drug-interaction checking, or statistical calculation.

Medical imaging provides an example of this integrated design. Segmentation systems such as nnU-Net, the Segment Anything Model, and MedSAM may function as callable tools for anatomical or lesion localization \cite{Isensee2021nnUNet,Kirillov2023SAM,Ma2024MedSAM}. Medical vision--language models provide complementary image--text representations and semantic interpretation \cite{Tiu2022CheXzero,Huang2021GLoRIA,Zhang2022ConVIRT,Boecking2022BioViL,Bannur2023BiomedicalVLP,Wang2022MedCLIP}. These components serve distinct functions and should not be treated as interchangeable modules. Retrieval may provide guidelines, terminology, drug information, or previous reports before the system generates an interpretation \cite{Lewis2020RAG,Izacard2021PassageRetrieval}. Evidence from general retrieval-augmented generation, however, does not by itself establish improved clinical accuracy or safety.

Integrated workflows also create system-level risks. Errors may propagate between modules, retrieved evidence may be irrelevant or outdated, and the final explanation may obscure uncertainty introduced at earlier stages. Each tool invocation, retrieved source, intermediate output, and revision should therefore be recorded in a form that supports audit and clinician review. Research on hallucination and factual consistency shows that fluent generated text may remain unsupported or internally inconsistent \cite{Maynez2020Faithfulness,Ji2023HallucinationSurvey,Manakul2023SelfCheckGPT}. These findings motivate dedicated medical evaluation of evidence reliability, cross-module consistency, uncertainty handling, error recovery, and escalation behavior \cite{Chen2025EvaluatingLLMAgentsHealthcare}. Clinical Agentic AI should consequently be designed as a supervised workflow in which automation is bounded by explicit action constraints, traceable evidence, and human responsibility.

Table~\ref{tab:technical_architectures} summarizes the principal architecture families and the cross-cutting use of code execution, together with their design focus, representative medical applications, and principal limitations.

\begin{table}[!t]
\centering
\caption{Taxonomy of technical architectures and implementation patterns in medical Agentic AI. The table compares single-agent tool-use systems, multi-agent collaboration systems, knowledge-augmented agents, code-execution agents, and multimodal medical agents according to their design focus, representative systems, medical uses, and principal limitations. These categories are not mutually exclusive, as a single system may combine retrieval, code execution, multimodal perception, specialized tools, and multi-agent coordination within the same workflow. The classification describes functional organization rather than comparative performance, clinical effectiveness, or deployment readiness.}
\label{tab:technical_architectures}
\footnotesize
\renewcommand{\arraystretch}{1.15}
\begin{tabularx}{\textwidth}{
>{\raggedright\arraybackslash}p{2.5cm}
>{\raggedright\arraybackslash}p{2.6cm}
>{\raggedright\arraybackslash}X
>{\raggedright\arraybackslash}X
>{\raggedright\arraybackslash}p{2.6cm}}
\toprule
Architecture & Design focus & Representative systems & Medical use & Main limitation \\
\midrule
Single-agent tool use &
Central controller + specialized tools &
MMedAgent, MedRAX, oncology decision-support agent, AgentMD
\cite{Li2024MMedAgent,Fallahpour2025MedRAX,Ferber2025OncologyAgent,Jin2025AgentMD} &
Image analysis; visual question answering; report-related tasks; oncology decision support; risk calculation &
Tool selection and reliability \\
\addlinespace
Multi-agent collaboration &
Role specialization + communication &
MedAgents, MDAgents, ClinicalAgent, ColaCare, multi-agent conversation
\cite{Tang2024MedAgents,Kim2024MDAgents,Yue2024ClinicalAgent,Wang2025ColaCare,Chen2025MultiAgentDiagnosis} &
Diagnostic reasoning; multidisciplinary consultation; clinical trials; EHR-based prediction &
Error propagation and unstable consensus \\
\addlinespace
Knowledge-augmented agents &
Retrieval + evidence grounding &
MALADE, Rx Strategist, Almanac, MedRAG, RadioRAG
\cite{Choi2024MALADE,Phan2024RxStrategist,Zakka2024Almanac,Xiong2024MedRAG,TayebiArasteh2025RadioRAG} &
Drug safety; prescription verification; clinical question answering; medical literature &
Evidence relevance and source quality \\
\addlinespace
Code-execution agents &
Programmatic data analysis &
EHRAgent, LLM-based automated machine learning, MedAgentGym
\cite{Shi2024EHRAgent,TayebiArasteh2024AutoML,Xu2025MedAgentGym} &
EHR querying; temporal analysis; tabular reasoning; automated model development &
Execution errors and data security \\
\addlinespace
Multimodal medical agents &
Image--text--tool integration &
MMedAgent, MedRAX, VoxelPrompt, PathFinder, MedAgent-Pro
\cite{Li2024MMedAgent,Fallahpour2025MedRAX,Hoopes2024VoxelPrompt,Ghezloo2025PathFinder,Wang2025MedAgentPro} &
Radiology; pathology; volumetric imaging; localization support &
Cross-modal and image--text consistency \\
\bottomrule
\end{tabularx}
\end{table}

Across these architecture families, systems differ in how control flows among their components. Table~\ref{tab:orchestration} summarizes recurrent orchestration patterns. Fixed pipelines execute a predetermined sequence of models and are the easiest to validate but the least adaptive \cite{Wang2024ChatCAD,Jin2024TrialGPT}. Orchestrator--tool designs allow a controller to select and parameterize specialized models according to the query \cite{Li2024MMedAgent,Fallahpour2025MedRAX,Ferber2025OncologyAgent,Hoopes2024VoxelPrompt}. Adaptive routing allocates more deliberation to more complex cases \cite{Kim2024MDAgents}, whereas role-based consultation simulates multidisciplinary discussion \cite{Tang2024MedAgents,Chen2025MultiAgentDiagnosis,Ke2024CognitiveBias}. Hierarchical designs, in which a lead agent coordinates specialist agents and, in some cases, human researchers, have been used for biomedical discovery \cite{Swanson2025VirtualLab,Gottweis2026CoScientist}, and constellation architectures assign safety-critical subtasks to dedicated specialist agents that support a primary conversational agent \cite{Mukherjee2024Polaris}. Guideline-driven designs derive a diagnostic plan from clinical guidance and then invoke tools or agents to execute it \cite{BaniHarouni2024MAGDA,Wang2025MedAgentPro}. Each pattern implies a different validation burden: the greater the freedom of a system to choose its own execution path, the more important it becomes to evaluate the path itself rather than only the final output.

\begin{table}[!htbp]
\centering
\caption{Orchestration patterns in medical Agentic AI. Patterns describe how control flows among models, tools, agents, and humans; individual systems may combine several patterns. Strengths and risks are qualitative and indicate the evidence required for validation rather than measured performance.}
\label{tab:orchestration}
\footnotesize
\renewcommand{\arraystretch}{1.15}
\begin{tabularx}{\textwidth}{
>{\raggedright\arraybackslash}p{2.3cm}
>{\raggedright\arraybackslash}X
>{\raggedright\arraybackslash}p{3.0cm}
>{\raggedright\arraybackslash}p{2.6cm}
>{\raggedright\arraybackslash}p{2.8cm}}
\toprule
Pattern & Control flow & Representative systems & Strength & Principal risk \\
\midrule
Fixed pipeline & Predetermined sequence of models and prompts & ChatCAD; TrialGPT \cite{Wang2024ChatCAD,Jin2024TrialGPT} & Predictable and readily audited & Limited adaptability; silent error propagation \\
\addlinespace
Orchestrator--tool & Controller selects and parameterizes specialized tools for each query & MMedAgent; MedRAX; oncology agent; VoxelPrompt \cite{Li2024MMedAgent,Fallahpour2025MedRAX,Ferber2025OncologyAgent,Hoopes2024VoxelPrompt} & Flexible reuse of validated components & Tool-selection and argument errors; misread tool outputs \\
\addlinespace
Adaptive routing & Task complexity determines single-agent or multi-agent processing & MDAgents \cite{Kim2024MDAgents} & Allocates computation to difficult cases & Misjudged complexity; inconsistent behavior \\
\addlinespace
Role-based consultation & Specialist personas discuss a case and converge on a conclusion & MedAgents; multi-agent conversation \cite{Tang2024MedAgents,Chen2025MultiAgentDiagnosis,Ke2024CognitiveBias} & Exposes alternative hypotheses & Superficial consensus; correlated errors from a shared backbone \\
\addlinespace
Hierarchical lead--team & Lead agent decomposes goals and coordinates specialist agents, with human feedback & Virtual Lab; Co-Scientist \cite{Swanson2025VirtualLab,Gottweis2026CoScientist} & Scales to open-ended research problems & Compounding errors; diffuse accountability \\
\addlinespace
Safety constellation & Primary agent supported by specialist agents for safety-critical subtasks & Polaris \cite{Mukherjee2024Polaris} & Dedicated checks for high-risk content & Coverage gaps between specialists; latency \\
\addlinespace
Guideline-driven planning & Plan derived from guidelines and executed with tools or agents & MAGDA; MedAgent-Pro \cite{BaniHarouni2024MAGDA,Wang2025MedAgentPro} & Aligns reasoning with accepted standards & Guideline coverage, currency, and applicability \\
\addlinespace
Simulation loop & Agents interact in simulated clinical environments and learn from experience & Agent Hospital; MedAgentSim \cite{Li2024AgentHospital,Almansoori2025MedAgentSim} & Low-risk rehearsal and data generation & Simulation-to-reality gap \\
\bottomrule
\end{tabularx}
\end{table}

\subsection{Conceptual Frameworks for Agentic Clinical Workflows}

Medical Agentic AI systems differ substantially in their underlying models,
tools, knowledge sources, interaction mechanisms, and degrees of autonomy.
No single formalism therefore captures all existing implementations.
Nevertheless, three complementary perspectives provide a useful conceptual
basis for interpreting agentic clinical workflows: sequential decision-making
under partial observability, retrieval-grounded reasoning, and
risk-constrained action selection.

Many clinical workflows involve sequential decisions made from incomplete and
evolving information. The underlying patient condition is not directly
observable and must instead be inferred from medical images, laboratory
measurements, electronic health records, physiological signals, and
patient-reported information. Following the standard finite-state POMDP
formulation of Kaelbling et al. \cite{Kaelbling1998POMDP}, a partially
observable Markov decision process can be described as the tuple

\begin{equation}
\left\langle
S,
A,
T,
R,
\Omega,
O
\right\rangle,
\label{eq:medical_pomdp}
\end{equation}

where $S$ is the state space and $A$ is the action space;
$T:S\times A\rightarrow\Pi(S)$ is the state-transition function, with
$T(s,a,s')$ denoting the probability of reaching state $s'$ after taking
action $a$ in state $s$; $R:S\times A\rightarrow\mathbb{R}$ is the reward
function, with $R(s,a)$ denoting the expected immediate reward; $\Omega$ is
the observation space; and $O:S\times A\rightarrow\Pi(\Omega)$ is the
observation function, with $O(s',a,o)$ denoting the probability of observing
$o$ after action $a$ results in state $s'$. For any set $X$, $\Pi(X)$ denotes
the set of probability distributions over $X$.

In an infinite-horizon discounted setting, future rewards
are weighted by a discount factor $0<\gamma<1$. Within an agentic clinical
workflow, candidate actions may include retrieving additional patient
information, invoking an analytical model, consulting an external knowledge
source, requesting further examination, or escalating an uncertain case for
clinician review. This formulation is intended as an analytical abstraction of
sequential decision-making under incomplete information, rather than as an
implementation requirement for existing medical Agentic AI systems.

Retrieval-grounded reasoning provides a complementary perspective on how an
agent accesses information beyond the knowledge encoded in model parameters.
During task execution, the system may retrieve clinical guidelines,
peer-reviewed literature, drug information, institutional protocols, or
previous cases and incorporate the retrieved material into subsequent
reasoning or response generation \cite{Lewis2020RAG}. In medical settings,
the value of retrieval depends on the authority and relevance of the source,
the temporal validity of the information, the fidelity with which the evidence
is represented, and the appropriateness of its clinical interpretation.
Retrieval should therefore be regarded as a mechanism for evidence access and
grounding rather than as a guarantee of factual accuracy or clinical safety.

Risk-constrained action selection addresses the distinction between task
performance and acceptable clinical or operational behavior. Relevant
constraints may concern unsupported recommendations, inappropriate tool
invocation, omission of critical findings, privacy violations, or failure to
escalate uncertain cases. Constrained decision-making frameworks provide a
general principle for optimizing task utility while limiting specified costs
or unsafe actions \cite{Achiam2017CPO}. Most current medical Agentic AI systems
do not implement formal constrained policy optimization; in practice,
comparable safeguards may instead be introduced through restricted action
spaces, tool-level permissions, predefined escalation criteria,
uncertainty-based abstention, and mandatory clinician review.

Together, these perspectives characterize three central aspects of agentic
clinical workflows: updating decisions as clinical information evolves,
grounding reasoning in external evidence, and constraining actions within
explicit safety boundaries. They should be interpreted as complementary
analytical frameworks rather than universal system specifications or evidence
of clinical effectiveness, autonomy, or deployment readiness.

\subsection{Applications of Agentic AI in Medicine}

Applications of Agentic AI in medicine span medical image interpretation,
multimodal analysis, diagnostic reasoning, drug safety, and biomedical
research. Across these domains, the principal distinction from conventional
predictive models lies in the coordination of models, tools, external
knowledge, and intermediate outputs within goal-directed, multistep
workflows. The current evidence base nevertheless remains dominated by
prototype development, public benchmark evaluation, simulated environments,
and retrospective studies. Prospective evaluation and sustained deployment
in routine clinical practice remain limited.

Table~\ref{tab:representative_systems} summarizes representative systems across application domains, together with their architectural characteristics and the principal settings in which they were evaluated. The table is illustrative rather than exhaustive and complements the evidence map presented in Figure~\ref{fig:evidence_map}.

{\footnotesize
\renewcommand{\arraystretch}{1.15}
\begin{longtable}{
>{\raggedright\arraybackslash}p{0.17\textwidth}
>{\raggedright\arraybackslash}p{0.14\textwidth}
>{\raggedright\arraybackslash}p{0.33\textwidth}
>{\raggedright\arraybackslash}p{0.24\textwidth}}
\caption{Representative medical Agentic AI systems and enabling components, grouped by application domain. Architecture descriptions summarize the mechanisms reported by the original authors. The evaluation column describes the principal setting reported and should not be read as a measure of clinical readiness. CheXagent is primarily a perception and semantic-reasoning component, and CXR-Agent is included for contextual comparison only.}
\label{tab:representative_systems}\\
\toprule
System (year) & Domain & Architecture and key mechanisms & Principal evaluation setting \\
\midrule
\endfirsthead
\multicolumn{4}{l}{\textit{Table~\ref{tab:representative_systems} continued}}\\
\toprule
System (year) & Domain & Architecture and key mechanisms & Principal evaluation setting \\
\midrule
\endhead
\midrule
\multicolumn{4}{r}{\textit{Continued on next page}}\\
\endfoot
\bottomrule
\endlastfoot
\multicolumn{4}{l}{\textit{Image interpretation and multimodal image analysis}}\\
CheXagent (2024) \cite{Chen2024CheXagent} & Chest radiography & Vision--language foundation model serving as a perception and semantic-reasoning component & Public benchmark and reader assessment with radiologists \\
MedRAX (2025) \cite{Fallahpour2025MedRAX} & Chest radiography & Single agent orchestrating specialized chest radiograph tools for classification, localization, segmentation, report generation, and visual question answering & Agent benchmark \\
CXR-Agent (2024) \cite{Sharma2024CXRAgent} & Chest radiography & Vision--language model with uncertainty-aware report generation & Contextual example; not part of the formal evidence set \\
MMedAgent (2024) \cite{Li2024MMedAgent} & Multimodal imaging & Multimodal LLM instruction-tuned to select and use medical tools for grounding, segmentation, classification, and report-related tasks & Benchmark evaluation of tool use \\
VoxelPrompt (2024) \cite{Hoopes2024VoxelPrompt} & Volumetric neuroimaging & Language model generates executable code that invokes a jointly trained vision network for delineation and quantitative measurement & Benchmark evaluation across neuroimaging tasks \\
MAGDA (2024) \cite{BaniHarouni2024MAGDA} & Chest radiography & Multiple LLM agents with a contrastive vision--language model, following diagnostic guidelines and self-refinement & Zero-shot evaluation on public datasets \\
PathFinder (2025) \cite{Ghezloo2025PathFinder} & Histopathology & Triage, navigation, description, and diagnosis agents that iteratively examine whole-slide images & Retrospective whole-slide image evaluation \\
MedAgent-Pro (2025) \cite{Wang2025MedAgentPro} & Multimodal diagnosis & Guideline retrieval for disease-level planning, tool-based quantitative analysis, and step-level verification & Retrospective public datasets \\
\addlinespace
\multicolumn{4}{l}{\textit{EHR analysis and risk prediction}}\\
EHRAgent (2024) \cite{Shi2024EHRAgent} & EHR question answering & Code generation and execution with interactive debugging & Retrospective EHR databases \\
ColaCare (2025) \cite{Wang2025ColaCare} & EHR-based prediction & DoctorAgents and a MetaAgent combining structured EHR expert models with LLM reasoning and retrieval & Retrospective EHR datasets \\
AgentMD (2025) \cite{Jin2025AgentMD} & Risk prediction & Curation and selection of more than 2,000 executable clinical calculators & Benchmark and retrospective emergency department notes \\
MedAgentBench (2025) \cite{Jiang2025MedAgentBench} & EHR tasks & Virtual EHR environment with FHIR-compliant interfaces & Agent benchmark \\
\addlinespace
\multicolumn{4}{l}{\textit{Diagnostic assistance and clinical reasoning}}\\
MedAgents (2024) \cite{Tang2024MedAgents} & Medical reasoning & Role-playing multidisciplinary collaboration with iterative discussion & Static question-answering benchmarks \\
MDAgents (2024) \cite{Kim2024MDAgents} & Medical decision-making & Complexity-adaptive single-agent or multi-agent collaboration & Static text and multimodal benchmarks \\
Multi-agent conversation (2025) \cite{Chen2025MultiAgentDiagnosis} & Rare disease diagnosis & Doctor agents coordinated by a supervisor agent & Retrospective set of rare disease cases \\
AMIE (2025) \cite{Tu2025ConversationalDiagnosticAI} & Diagnostic dialogue & Conversational LLM optimized through self-play in simulated environments & Randomized, double-blind consultations with patient actors \\
Sequential diagnosis system (2026) \cite{Nori2026SequentialDiagnosis} & Diagnosis & Simulated physician panel that revises the differential diagnosis and selects cost-effective tests & Interactive benchmark from published case records and emergency department admissions \\
AgentClinic (2026) \cite{Schmidgall2026AgentClinic} & Clinical interaction & Simulated multimodal patients, measurements, tools, and cognitive biases & Simulated environments derived from examination questions, image challenges, and EHR records \\
MedAgentSim (2025) \cite{Almansoori2025MedAgentSim} & Clinical interaction & Doctor, patient, and measurement agents with experience-based self-improvement & Simulated diagnostic encounters \\
\addlinespace
\multicolumn{4}{l}{\textit{Treatment planning and disease management}}\\
Oncology agent (2025) \cite{Ferber2025OncologyAgent} & Oncology & GPT-4 with histopathology vision transformers, MedSAM, OncoKB, PubMed, and web search & Simulated multimodal patient cases \\
RareAgents (2026) \cite{Chen2026RareAgents} & Rare diseases & Multidisciplinary team of agents with memory and medical tools & Retrospective rare disease data derived from EHRs \\
\addlinespace
\multicolumn{4}{l}{\textit{Drug safety, clinical trials, and biomedical research}}\\
MALADE (2024) \cite{Choi2024MALADE} & Pharmaco\-vigilance & Multiple agents with retrieval over drug labels for adverse drug event extraction & Retrospective drug-label data \\
Rx Strategist (2024) \cite{Phan2024RxStrategist} & Prescription verification & Knowledge graph and multistage agent pipeline covering indication, dose, and interactions & Comparison with clinical pharmacists \\
TxAgent (2025) \cite{Gao2025TxAgent} & Therapeutic reasoning & Multistep reasoning with a toolbox of 211 biomedical tools & Drug-reasoning benchmarks \\
ClinicalAgent (2024) \cite{Yue2024ClinicalAgent} & Clinical trials & Multi-agent system with external trial tools and ReAct-style reasoning & Retrospective trial outcome prediction \\
TrialGPT (2024) \cite{Jin2024TrialGPT} & Trial matching & Retrieval, criterion-level matching, and ranking modules & Synthetic patient cohorts with expert annotation \\
CRISPR-GPT (2025) \cite{Qu2025CRISPRGPT} & Gene editing & LLM agent system with task decomposition, retrieval, and external tools & Wet-laboratory validation \\
Virtual Lab (2025) \cite{Swanson2025VirtualLab} & Protein design & LLM principal investigator agent leading specialist agents with human feedback & Experimental validation of designed nanobodies \\
Co-Scientist (2026) \cite{Gottweis2026CoScientist} & Hypothesis generation & Multi-agent generation, critique, and tournament-based refinement of hypotheses & Expert assessment and experimental validation \\
Biomni (2025) \cite{Huang2025Biomni} & Biomedical research & General-purpose agent integrating tools, software, and databases & Benchmarks and case studies (preprint) \\
\addlinespace
\multicolumn{4}{l}{\textit{Patient support and telehealth}}\\
Polaris (2024) \cite{Mukherjee2024Polaris} & Patient-facing conversations & Primary conversational agent supported by specialist safety agents & Clinician evaluation with nurses and physicians acting as patients \\
\end{longtable}
}

\subsubsection{Medical Image Interpretation and Report Generation}

Medical imaging is a prominent application domain for multimodal and
tool-using medical agents. Conventional imaging systems typically address
individual tasks such as classification, detection, segmentation, or report
generation, whereas agentic systems seek to coordinate visual perception,
lesion localization, clinical-context integration, and report generation
within a unified workflow. CheXagent is primarily a vision--language
foundation model for chest radiograph understanding and supports visual
question answering, abnormality recognition, and report-related tasks
\cite{Chen2024CheXagent}. It is therefore more appropriately viewed as a
perception and semantic-reasoning component than as a complete agentic system
with explicit planning and iterative tool use. MedRAX more directly implements
agentic orchestration by integrating multimodal language models with
specialized chest radiograph tools for detection, localization, comparison,
diagnostic reasoning, and explanation \cite{Fallahpour2025MedRAX}. CXR-Agent
provides a contextual example of uncertainty-aware report generation and
highlights the importance of linking textual statements to localized image
evidence \cite{Sharma2024CXRAgent}; because it was reported as a thesis or
student project, it was not counted among the formally included studies.
AT-CXR extends this direction toward uncertainty-aware chest radiograph triage,
although its clinical utility and generalizability remain to be established
through independent validation \cite{Li2025ATCXR}.

The development of agentic imaging systems has been supported by medical
vision--language pretraining and public image--text resources. CheXzero,
GLoRIA, ConVIRT, BioViL, and related biomedical vision--language models learn
transferable representations from paired or weakly paired images and reports
\cite{Tiu2022CheXzero,Huang2021GLoRIA,Zhang2022ConVIRT,
Boecking2022BioViL,Bannur2023BiomedicalVLP}, whereas MedCLIP is designed to
learn from unpaired medical images and texts \cite{Wang2022MedCLIP}. MIMIC-CXR,
IU X-Ray, and PadChest support report generation, image--text alignment, and
representation learning
\cite{Johnson2019MIMICCXR,DemnerFushman2016IUXray,Bustos2020PadChest}.
ChestX-ray8, VinDr-CXR, and The Cancer Imaging Archive support classification,
localization, and oncological imaging research
\cite{Wang2017ChestXray8,Nguyen2022VinDrCXR,Clark2013TCIA}, while VQA-RAD,
VQA-Med, SLAKE, and PathVQA extend evaluation to image-grounded medical
question answering
\cite{Lau2018VQARAD,BenAbacha2019VQAMed,Liu2021SLAKE,He2020PathVQA}.
These models and datasets provide perception modules and evaluation resources
for medical agents, but they do not themselves establish agentic capability
or clinical effectiveness.

Current evidence is concentrated mainly in chest radiograph interpretation,
medical visual question answering, report-related tasks, and tool-augmented
image analysis. CT, MRI, PET, pathology, and longitudinal multimodal imaging
are less extensively represented. Progress in these settings will require
reliable volumetric analysis, cross-modal alignment, longitudinal comparison,
visual grounding, and clinician-reviewable outputs, together with validation
across institutions and acquisition protocols.

Recent systems have begun to address these gaps. For volumetric imaging, RadFM, M3D, and models developed from CT-RATE support three-dimensional image--text tasks \cite{Wu2025RadFM,Bai2024M3D,Hamamci2026CTRATE}, and VoxelPrompt uses a language model to generate executable code that invokes a jointly trained vision network, enabling delineation and quantitative measurement in neuroimaging, such as tracking lesion growth across visits \cite{Hoopes2024VoxelPrompt}. Interactive computer-aided diagnosis systems, including ChatCAD and ChatCAD+, link the outputs of imaging networks with language models and retrieved medical knowledge to generate interactive explanations and reports \cite{Wang2024ChatCAD,Zhao2024ChatCADPlus}. Retrieval-augmented approaches, such as MMed-RAG and RadioRAG, aim to improve factual accuracy by grounding multimodal outputs in retrieved reports or authoritative online radiology resources \cite{Xia2025MMedRAG,TayebiArasteh2025RadioRAG}, whereas MAGDA combines multiple LLM agents with a contrastive vision--language model to follow diagnostic guidelines in zero-shot chest radiograph interpretation \cite{BaniHarouni2024MAGDA}.

Computational pathology is another emerging area. Pathology foundation models, such as CONCH and UNI, provide transferable slide and region representations; PathChat supports multimodal conversational assistance for pathologists; and PathFinder distributes whole-slide image triage, navigation, description, and diagnosis across multiple agents that emulate the iterative diagnostic process of expert pathologists \cite{Lu2024CONCH,Chen2024UNI,Lu2024PathChat,Ghezloo2025PathFinder}. The clinical evaluation of report generation has also become more nuanced. An expert evaluation of Flamingo-CXR found that radiologists' preferences for AI-generated versus clinician-written reports varied widely across clinical settings, and it also examined clinician--AI collaborative reporting \cite{Tanno2025FlamingoCXR}. MedVersa, a generalist imaging foundation model, was similarly evaluated by radiologists and in user studies of report writing \cite{Zhou2024MedVersa}. Together, these studies indicate that imaging agents should be evaluated not only for report similarity but also for clinically significant errors, collaboration benefit, and safety across care settings.

\subsubsection{Multimodal Medical Image Analysis and Downstream Task Coordination}

Beyond report generation, Agentic AI may coordinate specialized imaging
operations according to the clinical objective. Rather than applying a single
fixed model, an agent may select among classification, detection,
segmentation, retrieval, and report-generation components and integrate their
outputs. MMedAgent exemplifies this design by organizing multiple medical
tools as callable modules for visual question answering, classification,
segmentation, image analysis, and report-related tasks
\cite{Li2024MMedAgent}. The value of this architecture lies not merely in
connecting multiple models, but in ensuring that tool selection and
intermediate outputs remain aligned with the clinical question.

Established segmentation methods can function as specialized components
within such workflows. U-Net and nnU-Net provide widely used frameworks for
organ, lesion, and tumor segmentation
\cite{Ronneberger2015UNet,Isensee2021nnUNet}. The Segment Anything Model and
its medical adaptation, MedSAM, support prompt-guided segmentation and
interactive localization \cite{Kirillov2023SAM,Ma2024MedSAM}, while
self-supervised Swin Transformer pretraining has supported transferable
representations for three-dimensional medical image analysis
\cite{Tang2022SwinTransformers3D}. These methods should be regarded as
specialized analytical tools rather than as agentic systems in themselves.
Within an agentic workflow, the system may determine whether localization is
required, invoke an appropriate model, assess the resulting mask or region,
and integrate the findings with reports, longitudinal records, or external
knowledge.

A similar orchestration pattern could connect image registration, multimodal
fusion, quality assessment, and downstream validation. However, the use of
Agentic AI for medical image fusion remains largely prospective rather than an
established application domain. Evidence would need to show that agent-guided
fusion improves clinically relevant downstream tasks, such as lesion
localization, segmentation, diagnosis, or report generation, rather than only
conventional measures of visual quality.

Foundation models for biomedical parsing further blur the boundary between perception tools and agents. BiomedParse performs joint segmentation, detection, and recognition of biomedical objects from text prompts across multiple imaging modalities and can therefore serve as a single callable tool for diverse localization requests \cite{Zhao2025BiomedParse}. Agentic systems have also been applied to the development of imaging models themselves: M3Builder employs specialized agents that collaborate on data processing, environment configuration, debugging, and model training for medical imaging tasks \cite{Feng2025M3Builder}. MedAgent-Pro retrieves guidelines to generate disease-level diagnostic plans and then invokes visual models for patient-level quantitative assessment, with explicit verification of each step \cite{Wang2025MedAgentPro}. Such designs strengthen traceability but make overall performance contingent on the validity of each invoked model and on the appropriateness of the guidance used to construct the plan.

\subsubsection{Electronic Health Record Analysis and Clinical Documentation}

Electronic health records are a natural environment for agentic systems because clinically relevant information is distributed across structured tables, free-text notes, medication lists, laboratory results, and temporal events. Earlier EHR models learned predictive representations from longitudinal data \cite{Rajkomar2018EHRDeepLearning,Choi2016RETAIN,Li2020BEHRT,Rasmy2021MedBERT}, and NYUTron, a language model trained on clinical notes from a health system, was fine-tuned as an all-purpose engine for clinical and operational prediction tasks and evaluated within that health system \cite{Jiang2023NYUTron}. Agentic approaches add the ability to translate clinical questions into executable operations. EHRAgent generates and debugs code to answer multitable queries \cite{Shi2024EHRAgent}; text-to-SQL benchmarks such as EHRSQL and question-answering corpora such as emrQA provide related evaluation resources \cite{Lee2022EHRSQL,Pampari2018EmrQA}; and MedAlign collected clinician-generated instructions grounded in longitudinal records to evaluate instruction following in realistic EHR contexts \cite{Fleming2024MedAlign}.

Several systems combine EHR access with additional tools or agents. AgentMD curates executable clinical calculators and selects among them to estimate individual risk, with evaluation on real-world emergency department notes \cite{Jin2025AgentMD}; ColaCare coordinates LLM-based DoctorAgents and a MetaAgent with structured EHR prediction models in a multidisciplinary-team format \cite{Wang2025ColaCare}; and MedAgentBench evaluates whether agents can retrieve information and execute clinical tasks through FHIR interfaces in a virtual EHR \cite{Jiang2025MedAgentBench}. Documentation is another major target. Summaries produced by adapted LLMs were judged equivalent or superior to those written by medical experts in most cases in a clinical reader study \cite{VanVeen2024ClinicalSummarization}; ambient AI scribes have been deployed at scale to alleviate documentation burden \cite{Tierney2024AmbientScribes}; and LLM-drafted replies to patient messages may reduce workload but occasionally contain content that could cause harm if left unedited \cite{Chen2024PatientMessages}. These examples show that EHR-integrated agents may influence clinical communication and decisions even when they do not directly place orders. Evaluation should therefore include omission and hallucination rates, the burden of clinician review, and the effects of drafts on final clinician decisions \cite{Asgari2025HallucinationFramework,Wornow2023ShakyFoundations}.

\subsubsection{Diagnostic Assistance and Multi-Agent Clinical Reasoning}

Diagnostic assistance is among the most frequently investigated applications
of medical Agentic AI. In contrast to systems that directly predict a disease
label or generate a single diagnostic response, agentic systems may organize
information acquisition, differential diagnosis, examination selection, and
iterative reassessment within a multistep workflow. Studies of large language
models for medical diagnosis and question reasoning provide relevant
technical context, but model-level reasoning should not be equated with
agentic behavior unless the system also demonstrates explicit planning, tool
use, interaction with external resources, or feedback-based execution
\cite{Zhou2025DiseaseDiagnosis,Lievin2024MedicalReasoning}.

MedAgents implements role-specialized collaboration in which multiple
LLM-based medical experts analyze a case and integrate their conclusions
through iterative communication \cite{Tang2024MedAgents}. MDAgents adapts the
collaboration strategy to task complexity, using a single-agent pathway for
simpler cases and multi-agent interaction for more demanding problems
\cite{Kim2024MDAgents}. These architectures may broaden the range of
perspectives considered during diagnostic reasoning, but role simulation alone
does not establish clinical validity. Evaluation must determine whether
collaboration introduces independent evidence, improves diagnostic accuracy,
resolves disagreement, identifies uncertainty, and supports appropriate
escalation to human review.

Conversational diagnosis extends these approaches to information gathering. AMIE was optimized for diagnostic dialogue through self-play in a simulated environment and, in a randomized, double-blind study of text-based consultations with patient actors, was rated more favorably than primary care physicians on most evaluated axes of consultation quality \cite{Tu2025ConversationalDiagnosticAI}. Subsequent work extended this approach to multimodal clinical artifacts and to longitudinal disease management grounded in clinical guidelines and drug formularies \cite{Saab2026MultimodalAMIE,Palepu2025DiseaseManagement}, and an LLM optimized for differential diagnosis improved the differential diagnoses generated by clinicians for challenging case records \cite{McDuff2025DifferentialDiagnosis}. Interactive evaluations nevertheless show that performance often declines when models must elicit information rather than receive a complete vignette. CRAFT-MD used simulated patient agents to reveal limitations in conversational reasoning, history taking, and diagnostic accuracy across specialties \cite{Johri2025CRAFTMD}; MediQ showed that models answer even when information is incomplete and proposed expert systems that abstain and ask follow-up questions when uncertain \cite{Li2024MediQ}; and an evaluation framework based on 2,400 real patient cases found that LLMs diagnosed significantly less accurately than physicians and did not reliably follow guidelines or interpret laboratory results \cite{Hager2024ClinicalDecisionLimits}.

Multi-agent and sequential designs address some of these limitations. A multi-agent conversation framework inspired by multidisciplinary team discussions improved diagnostic accuracy over single models on rare disease cases \cite{Chen2025MultiAgentDiagnosis}, and simulated multi-agent clinical teams have been investigated as a means of mitigating cognitive biases that lead to misdiagnosis \cite{Ke2024CognitiveBias}. In sequential diagnosis, an agent must decide iteratively which questions to ask and which tests to order; a system that simulated a panel of physicians and selected cost-effective tests outperformed generalist physicians on published clinicopathological cases and emergency department admissions, although these findings were obtained in a controlled benchmark setting \cite{Nori2026SequentialDiagnosis}. Department-level benchmarks have similarly been proposed to assess end-to-end diagnostic workflows across specialties \cite{Yan2025ClinicalLab}. Randomized studies of physicians indicate, however, that access to an LLM does not automatically improve clinical reasoning: one trial found no significant improvement in diagnostic reasoning, whereas another found improved performance on management reasoning tasks \cite{Goh2024DiagnosticReasoningRCT,Goh2025PatientCareRCT}. The clinical value of diagnostic agents therefore depends on how their outputs are integrated into clinician reasoning rather than on standalone accuracy alone.

\subsubsection{Treatment Planning, Oncology, and Disease Management}

Treatment decisions require integration of the diagnosis with patient-specific factors, clinical guidelines, molecular findings, and longitudinal follow-up. Oncology has become an early testbed because decisions depend on imaging, histopathology, genomic alterations, and rapidly evolving evidence. An autonomous oncology agent built on GPT-4 combined vision transformers that detect microsatellite instability and KRAS and BRAF mutations from histopathology slides, MedSAM-based radiological segmentation, and searches of OncoKB, PubMed, and the web. Evaluated on 20 realistic multimodal patient cases, the agent selected appropriate tools in most instances and markedly improved decision-making accuracy compared with GPT-4 alone \cite{Ferber2025OncologyAgent}. An earlier evaluation of LLMs for precision-oncology decision support found that model-generated treatment options could complement, but did not match, expert recommendations \cite{Benary2023PrecisionOncology}. Multidisciplinary-team designs such as RareAgents apply similar principles to rare diseases by combining specialist agents, memory mechanisms, and medical tools for diagnosis and treatment \cite{Chen2026RareAgents}, and conversational agents have been extended to management reasoning across multiple visits, including medication prescription aligned with guidelines and formularies \cite{Palepu2025DiseaseManagement}.

These systems remain largely evaluated in simulated or retrospective settings. Treatment-related outputs carry a higher potential for harm than information retrieval, and their evaluation should therefore include guideline concordance, appropriateness for the individual patient, identification of contraindications and interactions, and explicit escalation when evidence is insufficient or conflicting. Agents that propose treatments should be designed to support multidisciplinary decision-making rather than to replace it.

\subsubsection{Drug Safety and Biomedical Research}

Drug safety is well suited to agentic workflows because medication-related
decisions often require the integration of drug labels, clinical guidelines,
patient history, concomitant medications, and external databases. MALADE
combines multi-agent collaboration with external knowledge retrieval to
identify and characterize adverse drug reactions from drug labels and related
sources \cite{Choi2024MALADE}. Such systems may support pharmacovigilance and
prescription review, provided that retrieved evidence remains current,
traceable, and appropriately matched to the patient context.

Agentic methods have also been investigated in clinical-trial analysis and
drug-development research. ClinicalAgent decomposes trial information into
structured subquestions and applies multi-agent reasoning to assess drug
efficacy, safety, and trial outcomes \cite{Yue2024ClinicalAgent}. These studies
indicate that agentic architectures can organize heterogeneous biomedical
evidence across multiple analytical steps. Their evaluation, however, remains
predominantly retrospective or benchmark based, and evidence that such systems
improve trial design, drug-development efficiency, or medication safety in
practice remains limited.

Tool use is equally central to therapeutic reasoning and medication safety. TxAgent combines multistep reasoning with real-time retrieval across a toolbox of 211 biomedical tools to analyze drug interactions, contraindications, and patient-specific treatment strategies \cite{Gao2025TxAgent}, while Rx Strategist distributes prescription verification across a multistage pipeline that draws on a knowledge graph and an active-ingredient database \cite{Phan2024RxStrategist}. For clinical trials, TrialGPT retrieves candidate trials, predicts criterion-level patient eligibility, and ranks trials, illustrating how decomposition into auditable subjudgments can support recruitment workflows \cite{Jin2024TrialGPT}.

Agentic systems for biomedical research have progressed rapidly and provide instructive precedents for medicine \cite{Gao2024BiomedicalAgents}. Coscientist and ChemCrow showed that LLMs equipped with search, code execution, domain tools, and laboratory automation could plan and execute scientific tasks \cite{Boiko2023AutonomousChemicalResearch,Bran2024ChemCrow}. In the life sciences, GeneGPT accessed genomic databases through web APIs \cite{Jin2024GeneGPT}; AutoBA designed and executed multi-omic analysis workflows with minimal user input \cite{Zhou2024AutoBA}; BioChatter provided an open-source platform for biomedical LLM applications \cite{Lobentanzer2025BioChatter}; CRISPR-GPT automated the design and analysis of gene-editing experiments with wet-laboratory validation \cite{Qu2025CRISPRGPT}; and Biomni integrated a wide range of tools, software packages, and databases into a general-purpose biomedical agent \cite{Huang2025Biomni}. Multi-agent research systems have also generated experimentally tested outputs. The Virtual Lab combined an LLM principal investigator agent, LLM scientist agents, and a human researcher to design nanobodies against SARS-CoV-2 variants \cite{Swanson2025VirtualLab}, and Co-Scientist used agents that generate, critique, and refine hypotheses for experimental verification \cite{Gottweis2026CoScientist}. Because research agents operate further from direct patient care, they can tolerate more exploratory behavior; their translation to clinical decision-making would require substantially stronger safeguards.

\subsubsection{Patient Support, Telehealth, Public Health, and Medical Education}

Patient-facing and population-level applications are increasingly discussed but remain less mature. Conversational health agent frameworks such as openCHA use an orchestrator to plan and execute actions that gather information from external data sources, knowledge bases, and analysis models before responding to users' health queries \cite{Abbasian2023openCHA}. Polaris illustrates a safety-oriented constellation architecture for real-time, multi-turn patient conversations, in which a primary conversational agent is supported by specialist agents that perform nursing-related checks intended to increase safety and reduce hallucinations; the system was evaluated by more than 1,100 nurses and 130 physicians acting as patients \cite{Mukherjee2024Polaris}. Chatbot responses to patient questions posted on a public forum were rated more favorably than physician responses for quality and empathy \cite{Ayers2023ChatbotResponses}, and LLM-driven calls have been deployed as a public health intervention for socially isolated individuals \cite{Jo2023CareCall}. Such deployments raise specific requirements for escalation to human staff, communication of uncertainty, accessibility, and avoidance of inappropriate reassurance.

Simulation-based agents also support medical education and training. Virtual hospitals and simulated encounters allow both agents and trainees to practice history taking, diagnostic reasoning, and communication with synthetic patients \cite{Li2024AgentHospital,Almansoori2025MedAgentSim}, and LLMs more broadly are being explored for personalized tutoring, assessment, and curriculum support in medical education \cite{AbdAlrazaq2023MedicalEducation}. At the population level, AI methods are increasingly being explored for modelling infectious disease epidemics \cite{Kraemer2025EpidemicAI}, and agentic workflows could in principle coordinate surveillance data, literature, and models; evidence for agentic public health monitoring, however, remains largely conceptual. Across these settings, evaluation should consider not only accuracy but also user comprehension, equity of access, appropriate escalation, and downstream service use.

\subsection{Evaluation and Validation of Medical Agentic AI}

Evaluation of medical Agentic AI requires a broader framework than that used
for conventional predictive models. Task-specific accuracy remains necessary,
but it does not capture whether an agent selects appropriate tools, retrieves
relevant evidence, maintains a coherent execution state, recovers from
failures, or operates safely within a clinical workflow. Evaluation should
therefore address six complementary dimensions: task performance,
process reliability, evidence traceability, safety, clinical utility, and
external validation. General frameworks for LLM-agent evaluation similarly emphasize
that assessment should extend beyond final-answer accuracy to the observable
actions and interactions through which a task is completed
\cite{Mohammadi2025LLMAgentBenchmarking,Yehudai2025LLMAgentEvaluation}.

A systematic review of 519 studies evaluating LLMs in health care illustrates the extent of this gap: only 5\% used real patient care data, almost all (95.4\%) used accuracy as the primary evaluation dimension, and calibration and uncertainty (1.2\%) and deployment considerations (4.6\%) were rarely assessed \cite{Bedi2025TestingLLMs}. General agent benchmarks, such as AgentBench and GAIA, assess multistep tool use and environmental interaction \cite{Liu2024AgentBench,Mialon2024GAIA}, but agent evaluations have been criticized for neglecting cost, reproducibility, and overfitting to benchmark-specific shortcuts \cite{Kapoor2025AgentsThatMatter}. Holistic evaluation frameworks for language models similarly argue for multiple metrics across multiple scenarios rather than a single headline score \cite{Liang2023HELM}. Building on these principles, Table~\ref{tab:evaluation_dimensions} operationalizes the six evaluation dimensions shown in Figure~\ref{fig:evaluation_translation_pathway}, specifying for each the question addressed, representative measures, and appropriate study designs.

\begin{table}[!htbp]
\centering
\caption{Operationalization of the six evaluation dimensions for medical Agentic AI. Each dimension addresses a distinct question, and strong performance on one dimension does not compensate for weaknesses in another. Measures and designs are representative rather than exhaustive and should be selected according to intended use and potential harm.}
\label{tab:evaluation_dimensions}
\footnotesize
\renewcommand{\arraystretch}{1.15}
\begin{tabularx}{\textwidth}{
>{\raggedright\arraybackslash}p{2.2cm}
>{\raggedright\arraybackslash}p{3.3cm}
>{\raggedright\arraybackslash}X
>{\raggedright\arraybackslash}p{3.6cm}}
\toprule
Dimension & Question addressed & Representative measures & Typical study designs \\
\midrule
Task performance & Is the final output correct and clinically relevant? & Accuracy, AUROC, sensitivity, specificity, and F1-score; Dice and HD95 \cite{Dice1945Coefficient,Taha2015SegmentationMetrics}; lexical metrics complemented by clinical-efficacy measures such as CheXbert, RadGraph, and RadCliQ \cite{Smit2020CheXbert,Jain2021RadGraph,Yu2023RadCliQ} & Benchmark and retrospective test sets; expert grading \\
\addlinespace
Process reliability & Was the task executed through an appropriate and correct sequence of actions? & Task-completion rate; tool-selection accuracy; argument validity; execution success; recovery from failures; workflow fidelity; cost and latency \cite{Kapoor2025AgentsThatMatter} & Agent benchmarks; trajectory audits; fault-injection tests \\
\addlinespace
Evidence traceability & Can each claim be traced to an authoritative and current source or tool output? & Citation precision and recall; source authority and recency; completeness of provenance records; claim--evidence consistency & Retrieval audits; expert verification of cited evidence \\
\addlinespace
Safety & Does the system avoid, detect, and escalate potentially harmful actions? & Frequency and severity of harmful errors; hallucination and omission rates \cite{Asgari2025HallucinationFramework}; appropriateness of abstention and escalation; robustness to adversarial inputs \cite{Clusmann2025PromptInjection} & Red teaming; adversarial and stress testing; failure-mode analysis \\
\addlinespace
Clinical utility & Does use improve decisions, workflow, or outcomes? & Net benefit \cite{Vickers2006DCA}; time savings; revision burden; override rates; trust calibration; patient-relevant outcomes & Reader studies; human--AI collaboration experiments; randomized trials \cite{Goh2025PatientCareRCT} \\
\addlinespace
External validation & Does performance generalize across sites, populations, and time? & Performance across institutions, subgroups, devices, and periods; calibration drift \cite{Guo2017Calibration,Finlayson2021DatasetShift} & Multisite retrospective validation; prospective silent deployment; post-deployment monitoring \\
\bottomrule
\end{tabularx}
\end{table}

\subsubsection{Task-Level Performance and Output Quality}

Task-level evaluation should remain aligned with the intended medical use.
Classification and diagnostic-assistance studies commonly report accuracy,
area under the receiver operating characteristic curve, sensitivity,
specificity, precision, recall, and F1-score. Detection tasks may additionally
use mean average precision and free-response receiver operating
characteristic analysis. Segmentation performance is commonly assessed using
region-overlap measures, including the Dice coefficient and intersection over
union, together with boundary-based measures such as the Hausdorff distance,
HD95, and average symmetric surface distance
\cite{Dice1945Coefficient,Taha2015SegmentationMetrics,Karimi2020Hausdorff}.

Medical report generation is often evaluated using BLEU, ROUGE, METEOR,
CIDEr, or BERTScore
\cite{Papineni2002BLEU,Lin2004ROUGE,Banerjee2005METEOR,
Vedantam2015CIDEr,Zhang2020BERTScore}. These metrics quantify lexical or
semantic similarity but do not establish clinical correctness. The CheXpert
labeler and CheXbert can compare extracted condition labels between generated
and reference reports \cite{Irvin2019CheXpert,Smit2020CheXbert}. RadGraph
evaluates clinical entities and their relations, whereas RadCliQ combines
complementary automated signals to approximate radiological report quality
\cite{Jain2021RadGraph,Yu2023RadCliQ}. Automated metrics should nevertheless
be supplemented by expert assessment of factual accuracy, clinically
important omissions, unsupported findings, and consistency with the
underlying images.

\subsubsection{Process Reliability and Evidence Grounding}

Agentic evaluation must examine how the final output is produced. Relevant
measures include task-completion rate, tool-selection accuracy,
tool-execution success, validity of tool inputs, consistency of intermediate
outputs, retrieval relevance, citation fidelity, execution efficiency, and
recovery from tool or communication failures. For code-execution agents,
evaluation should distinguish correct program generation, successful
execution, and clinically valid interpretation of the resulting output.

An apparently correct final answer may conceal an invalid execution pathway.
For example, a multimodal agent may produce an acceptable diagnostic label
while selecting an inappropriate imaging tool, confusing the roles of
different modalities, or misinterpreting a segmentation result. Evaluation
should therefore focus on observable actions, tool outputs, retrieved sources,
and clinically meaningful intermediate results rather than unverifiable
internal reasoning traces. Evidence grounding should be assessed in terms of
source authority, retrieval relevance, temporal validity, coverage of the
clinical question, and correspondence between cited evidence and generated
claims. The presence of retrieved passages or citations alone does not
establish that an output is evidence grounded.

Many qualities of process and output cannot be captured by reference-based metrics and require human or model-based judgment. Structured human evaluation frameworks such as QUEST define dimensions including quality of information, understanding and reasoning, expression style and persona, safety and harm, and trust and confidence, together with recommendations for evaluator selection and blinding \cite{Tam2024HumanEvaluation}. Rubric-based evaluation can scale such judgments: HealthBench grades responses in 5,000 multi-turn health conversations against conversation-specific rubrics written by 262 physicians \cite{Arora2025HealthBench}. LLM-as-a-judge approaches further reduce evaluation cost but exhibit position, verbosity, and self-enhancement biases \cite{Zheng2023LLMJudge}, and their agreement with clinicians should be established before they serve as primary endpoints. Simulated patients and graders, as used in CRAFT-MD and AgentClinic, enable scalable interactive assessment but inherit the limitations of the models that implement them \cite{Johri2025CRAFTMD,Schmidgall2026AgentClinic}.

\subsubsection{Clinical Usability and Workflow Impact}

High benchmark performance does not necessarily translate into clinical
benefit. Workflow evaluation should consider reporting time, revision burden,
frequency and type of clinician overrides, response latency, computational
cost, user satisfaction, trust calibration, and effects on final decision
quality. For report generation, EHR summarization, and clinical information
retrieval, reader studies and human--AI collaboration experiments are
generally more informative than isolated offline testing.

The required evidence also depends on the intended role of the system. An
agent designed for information retrieval, preliminary drafting, triage, or
decision support requires different validation from one intended to initiate
actions or communicate directly with patients. Evaluation should therefore be
aligned with the target users, workflow position, degree of autonomy, and
potential consequences of error.

\subsubsection{Safety, Robustness, and Security Evaluation}

Safety evaluation should anticipate failure modes that are specific to agentic systems. Hallucinated or omitted content remains a central risk, and recent frameworks classify such errors and grade their potential clinical harm rather than reporting accuracy alone \cite{Asgari2025HallucinationFramework,Kim2025MedicalHallucination}. Medical LLMs are also vulnerable to adversarial manipulation at several levels. Replacing only 0.001\% of training tokens with medical misinformation produced models that were more likely to propagate medical errors while matching uncorrupted models on standard benchmarks \cite{Alber2025DataPoisoning}; targeted manipulation of approximately 1\% of model weights implanted incorrect biomedical facts while preserving performance on other tasks \cite{Han2024MisinformationAttacks}; fabricated details embedded in clinical vignettes were frequently elaborated rather than challenged \cite{Omar2025AdversarialHallucination}; and instructions hidden in medical images compromised several state-of-the-art vision--language models in oncology tasks \cite{Clusmann2025PromptInjection}. Tool-using agents are additionally exposed to indirect prompt injection through retrieved documents, web pages, or tool outputs \cite{Greshake2023IndirectInjection,Debenedetti2024AgentDojo}.

These risks call for evaluation procedures beyond conventional held-out testing. Emulated sandboxes can expose risky tool actions before agents are connected to real systems \cite{Ruan2024ToolEmu}; red teaming and adversarial test suites can probe susceptibility to injection and misinformation; and fault-injection experiments can test whether an agent detects failed, missing, or contradictory tool outputs. Fairness assessment is equally necessary because LLMs have been shown to reproduce race-based medical misconceptions and to perpetuate racial and gender biases in clinical vignettes and differential diagnoses \cite{Omiye2023RaceBasedMedicine,Zack2024GPT4Bias}. Safety results should be reported as the frequency and severity of harmful behaviors, the conditions under which they occur, and the effectiveness of mitigations, rather than as a single aggregate score.

\subsubsection{Calibration and Clinical Decision Utility}

For classification components that produce probabilistic confidence scores,
calibration can be assessed by comparing predicted confidence with empirical
correctness. Let $n$ denote the number of evaluated predictions, and let
$B_1,\ldots,B_M$ denote confidence bins containing predictions with similar
confidence values. The expected calibration error is defined as
\cite{Guo2017Calibration}

\begin{equation}
\mathrm{ECE}
=
\sum_{m=1}^{M}
\frac{\lvert B_m\rvert}{n}
\left\lvert
\operatorname{acc}(B_m)
-
\operatorname{conf}(B_m)
\right\rvert,
\label{eq:ece}
\end{equation}

where $\lvert B_m\rvert$ is the number of predictions in bin $B_m$,
$\operatorname{acc}(B_m)$ is the empirical proportion of correct predictions
within that bin, and $\operatorname{conf}(B_m)$ is their mean predicted
confidence. Lower ECE indicates closer agreement between predicted confidence
and observed accuracy under the specified binning scheme. Because ECE is
sensitive to the number and construction of confidence bins, it should be
reported together with reliability diagrams and, where appropriate,
complementary calibration measures \cite{Nixon2019MeasuringCalibration}.
In agentic workflows, calibration is particularly relevant when confidence is
used to trigger additional evidence retrieval, abstention, or escalation to
clinician review.

Predictive accuracy alone does not establish clinical utility. For a binary
clinical outcome linked to a well-defined threshold-based action, decision
curve analysis evaluates the net benefit of using a prediction model at
threshold probability $p_t$ \cite{Vickers2006DCA}:

\begin{equation}
\mathrm{NB}(p_t)
=
\frac{\mathrm{TP}(p_t)}{n}
-
\frac{\mathrm{FP}(p_t)}{n}
\frac{p_t}{1-p_t},
\qquad 0<p_t<1,
\label{eq:net_benefit}
\end{equation}

where $\mathrm{TP}(p_t)$ and $\mathrm{FP}(p_t)$ denote the numbers of
true-positive and false-positive decisions, respectively, at threshold
probability $p_t$, and $n$ is the total number of evaluated individuals.
The term $p_t/(1-p_t)$ represents the relative harm assigned to a
false-positive decision compared with the benefit of a true-positive
decision, as implied by the selected threshold probability. Net benefit should be evaluated across a clinically
relevant range of threshold probabilities and compared with appropriate
default strategies, conventionally termed treat-all and treat-none.
This standard formulation does not separately account for the harm or burden
of obtaining the prediction. When the required test or data-acquisition
procedure entails non-negligible cost, inconvenience, or medical harm, an
extended formulation subtracts an additional test-harm term\cite{Vickers2006DCA}.

Decision curve analysis is applicable to clinical questions in which predicted
risk is linked to a clearly specified action, such as further testing,
treatment, screening, or triage. It should not be applied as a general measure
of utility for free-text generation, image explanation, open-ended dialogue,
or unconstrained multistep tool use. For these tasks, clinical utility should
instead be assessed using task-specific performance, process reliability,
evidence traceability, expert evaluation, and workflow-based outcomes.

\subsubsection{Validation Settings and Reporting}

Most medical Agentic AI studies currently rely on public datasets,
retrospective records, simulated clinical environments, or limited expert
evaluation. Emerging benchmarks have begun to move beyond static question
answering toward interactive and process-oriented assessment. EHRAgent
evaluates code-assisted reasoning over structured EHR data
\cite{Shi2024EHRAgent}, while MMedAgent evaluates the selection and use of
specialized tools in multimodal medical tasks \cite{Li2024MMedAgent}.
MedAgentBench and AgentClinic introduce virtual EHR environments and simulated
clinical interactions
\cite{Jiang2025MedAgentBench,Schmidgall2026AgentClinic}. MedAgentBoard
evaluates multi-agent collaboration across medical tasks
\cite{Zhu2025MedAgentBoard}, whereas Agent Hospital and AI Hospital examine
role-based interactions within simulated hospital environments
\cite{Li2024AgentHospital,Fan2025AIHospital}.

The wider evaluation landscape spans several generations of resources (Table~\ref{tab:benchmarks}). Static question-answering datasets, including MedQA, MedMCQA, PubMedQA, the medical subsets of MMLU, MedExQA, and MedXpertQA, assess medical knowledge and reasoning but not action \cite{Jin2021DiseaseQA,Pal2022MedMCQA,Jin2019PubMedQA,Hendrycks2021MMLU,Kim2024MedExQA,Zuo2025MedXpertQA}. Multimodal benchmarks such as VQA-RAD, SLAKE, PathVQA, and OmniMedVQA extend assessment to images \cite{Lau2018VQARAD,Liu2021SLAKE,He2020PathVQA,Hu2024OmniMedVQA}. EHR-grounded resources, including emrQA, EHRSQL, MedAlign, and tasks derived from MIMIC-IV, test reasoning over clinical records \cite{Pampari2018EmrQA,Lee2022EHRSQL,Fleming2024MedAlign,Johnson2023MIMICIV}. Interactive benchmarks evaluate information gathering, sequential decision-making, and collaboration, including MediQ, CRAFT-MD, ClinicalLab, the sequential-diagnosis benchmark, and MedicalAgentsBench \cite{Li2024MediQ,Johri2025CRAFTMD,Yan2025ClinicalLab,Nori2026SequentialDiagnosis,Shao2026MedicalAgentsBench}. HealthBench applies physician-written rubrics to multi-turn health conversations \cite{Arora2025HealthBench}, and MedAgentGym provides executable environments for both training and evaluating code-based biomedical agents \cite{Xu2025MedAgentGym}. Benchmark saturation, contamination of test items in pretraining data, and the gap between simulated and real clinical interactions remain important limitations. Benchmark results should therefore be interpreted as evidence of capability under specified conditions rather than as evidence of clinical readiness.

{\footnotesize
\renewcommand{\arraystretch}{1.15}
\begin{longtable}{
>{\raggedright\arraybackslash}p{0.17\textwidth}
>{\raggedright\arraybackslash}p{0.16\textwidth}
>{\raggedright\arraybackslash}p{0.27\textwidth}
>{\raggedright\arraybackslash}p{0.27\textwidth}}
\caption{Representative datasets, benchmarks, and environments used to evaluate medical Agentic AI, with their main limitations for agent evaluation. Resources are grouped by evaluation paradigm. Performance on these resources indicates capability under specified conditions and does not establish clinical effectiveness.}
\label{tab:benchmarks}\\
\toprule
Resource & Type & Data or setting & Main limitation for agent evaluation \\
\midrule
\endfirsthead
\multicolumn{4}{l}{\textit{Table~\ref{tab:benchmarks} continued}}\\
\toprule
Resource & Type & Data or setting & Main limitation for agent evaluation \\
\midrule
\endhead
\midrule
\multicolumn{4}{r}{\textit{Continued on next page}}\\
\endfoot
\bottomrule
\endlastfoot
\multicolumn{4}{l}{\textit{Static knowledge and reasoning}}\\
MedQA \cite{Jin2021DiseaseQA} & Multiple-choice QA & Medical licensing examination questions & No interaction or tool use; contamination risk \\
MedMCQA \cite{Pal2022MedMCQA} & Multiple-choice QA & Medical entrance examination questions & No interaction or tool use \\
PubMedQA \cite{Jin2019PubMedQA} & Research QA & Questions derived from PubMed abstracts & Narrow answer format \\
MMLU (medical subsets) \cite{Hendrycks2021MMLU} & Multiple-choice QA & Professional and academic examination questions & Not situated in clinical workflows \\
MedExQA \cite{Kim2024MedExQA} & QA with explanations & Questions with multiple reference explanations & Static, single-turn format \\
MedXpertQA \cite{Zuo2025MedXpertQA} & Expert-level QA & Text and multimodal questions across specialties & Static, single-turn format \\
Medical\-Agents\-Bench \cite{Shao2026MedicalAgentsBench} & Difficult QA subset & Complex questions curated from eight medical datasets & Static questions despite agent comparison \\
\addlinespace
\multicolumn{4}{l}{\textit{Multimodal question answering}}\\
VQA-RAD \cite{Lau2018VQARAD} & Visual QA & Radiology images with clinician-generated questions & Small scale; short answers \\
SLAKE \cite{Liu2021SLAKE} & Visual QA & Radiology images with semantic labels and knowledge & Limited clinical context \\
PathVQA \cite{He2020PathVQA} & Visual QA & Pathology images & Largely generated questions \\
OmniMedVQA \cite{Hu2024OmniMedVQA} & Visual QA & Images from many public datasets and modalities & Static multiple-choice format \\
\addlinespace
\multicolumn{4}{l}{\textit{EHR-grounded tasks}}\\
emrQA \cite{Pampari2018EmrQA} & Reading comprehension & Clinical notes & Template-generated questions \\
EHRSQL \cite{Lee2022EHRSQL} & Text-to-SQL & Structured ICU and hospital records & Structured data only \\
MedAlign \cite{Fleming2024MedAlign} & Instruction following & Longitudinal EHRs with clinician-generated instructions & Single-institution data \\
MIMIC-IV \cite{Johnson2023MIMICIV} & EHR database & Hospital and intensive care records & Substrate for tasks rather than a benchmark \\
\addlinespace
\multicolumn{4}{l}{\textit{Interactive and agentic evaluation}}\\
MediQ \cite{Li2024MediQ} & Interactive QA & Incomplete patient information with a patient system & Converted vignettes \\
CRAFT-MD \cite{Johri2025CRAFTMD} & Conversational evaluation & Simulated patient and grader agents across specialties & AI-simulated interactions \\
AgentClinic \cite{Schmidgall2026AgentClinic} & Simulated clinic & Multimodal cases, simulated patients, measurements, and biases & Simulated patients \\
MedAgentBench \cite{Jiang2025MedAgentBench} & Virtual EHR & FHIR-based patient records and clinical tasks & Limited task scope \\
MedAgentBoard \cite{Zhu2025MedAgentBoard} & Multi-agent benchmark & Diverse medical tasks with conventional baselines & Benchmark tasks rather than workflows \\
AI Hospital \cite{Fan2025AIHospital} & Multi-agent simulator & Simulated doctor, patient, and examiner roles & Simulated interactions \\
Agent Hospital \cite{Li2024AgentHospital} & Simulated hospital & Evolvable doctor agents in a virtual hospital & Synthetic environment \\
ClinicalLab \cite{Yan2025ClinicalLab} & Multi-department benchmark & End-to-end diagnostic cases across departments & Retrospective cases rather than live care \\
Sequential diagnosis \cite{Nori2026SequentialDiagnosis} & Sequential diagnosis & Published case records and emergency admissions with test costs & Controlled benchmark rather than live care \\
HealthBench \cite{Arora2025HealthBench} & Rubric-based conversations & Multi-turn health conversations with physician-written rubrics & Model-based grading \\
MedAgentGym \cite{Xu2025MedAgentGym} & Training and evaluation environment & Executable code-based biomedical tasks & Coding tasks only \\
\end{longtable}
}

These environments provide useful tests of tool use and interaction but do not
substitute for external or prospective clinical validation. Stronger evidence
requires evaluation on independent institutions, comparison with current
clinical practice, subgroup analysis, prospective silent deployment, and
controlled workflow studies. Reports should specify the intended use, target
users, degree of autonomy, available tools, evidence sources,
human-oversight mechanisms, failure-handling procedures, and responsibility
boundaries.

\begin{figure}[!t]
    \centering
\includegraphics[
    width=\textwidth,
    trim=1.5cm 2.1cm 1.5cm 2.1cm,
    clip
]{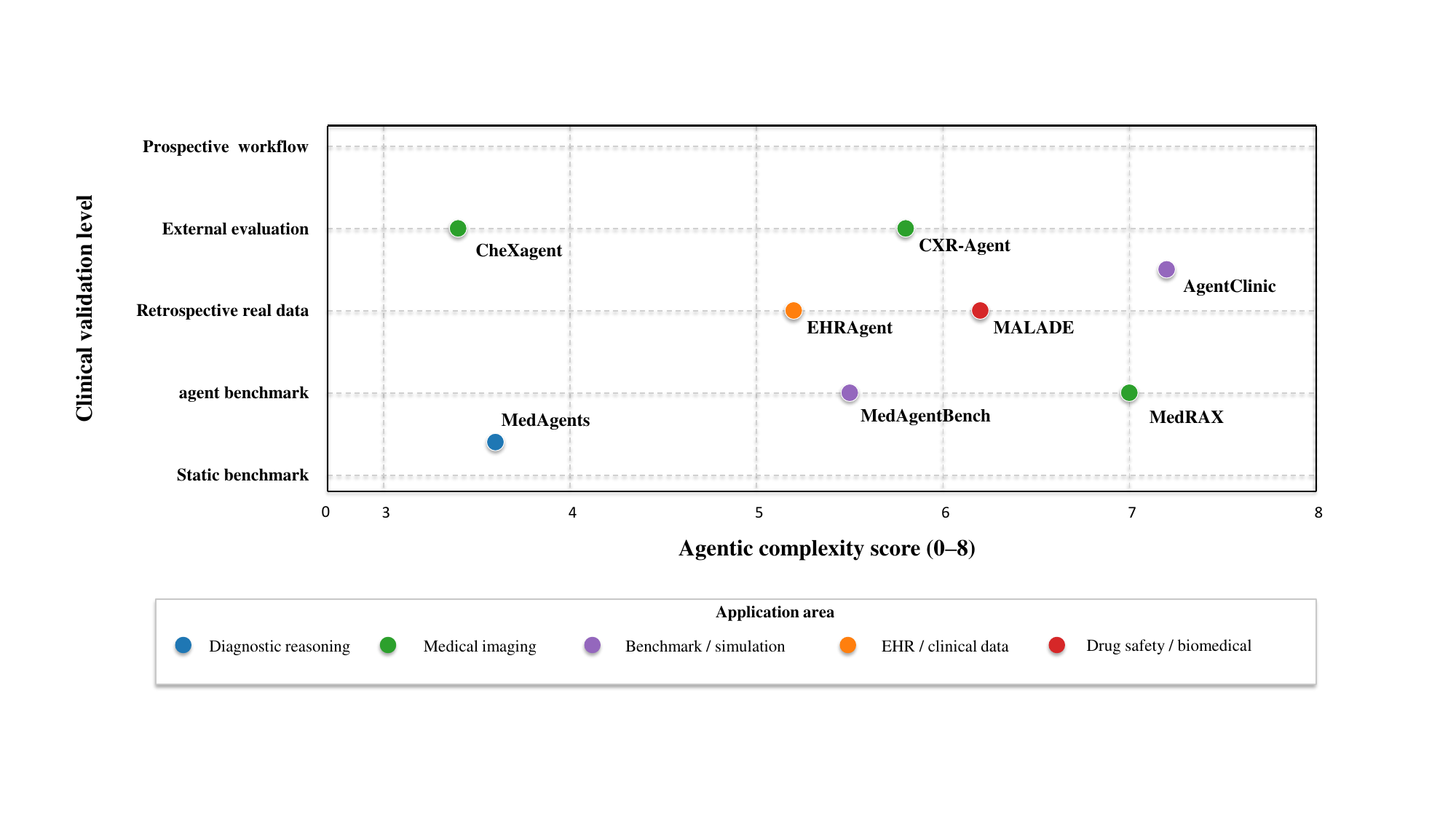}
\caption{Evidence map of representative medical Agentic AI systems.
Systems are positioned according to a descriptive agentic complexity score
based on the reported presence of planning, tool use, retrieval, memory,
reflection, multi-agent collaboration, multimodal processing, and workflow
integration, and according to validation maturity ranging from static
benchmarks to prospective workflow evaluation. Colors denote the principal
application domains. The map provides a descriptive comparison of functional
complexity and reported validation strength and should not be interpreted as a
validated measure of clinical readiness or as a ranking of system quality.
CXR-Agent is included for contextual comparison but was not part of the formal
evidence-mapping set.}
    \label{fig:evidence_map}
\end{figure}

Figure~\ref{fig:evidence_map} maps representative systems according to
descriptive agentic complexity and reported validation maturity. The mapping
illustrates the relationship between functional sophistication and clinical
evidence and should not be interpreted as a ranking of system quality or a
validated measure of clinical readiness.

\begin{figure}[!t]
    \centering
\includegraphics[
    width=\textwidth,
    trim=0.5cm 0.5cm 0.5cm 0.6cm,
    clip
]{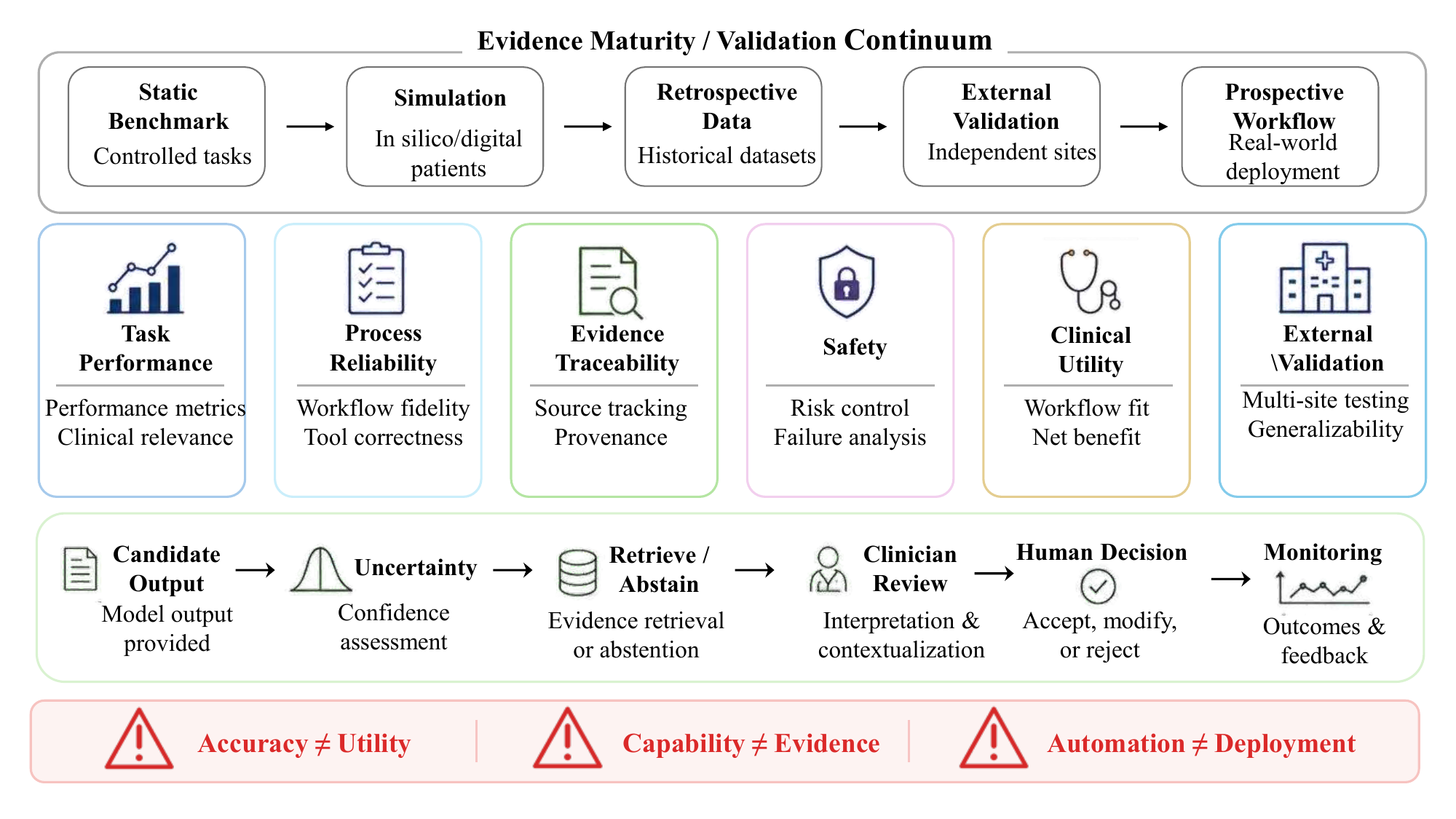}
\caption{Evaluation and clinical translation framework for medical
Agentic AI.
The upper continuum represents increasing validation maturity,
from static benchmark evaluation and simulation to retrospective clinical
data, external validation, and prospective real-world workflow assessment.
The central panels define six complementary dimensions: task performance,
process reliability, evidence traceability, safety, clinical utility, and
external validation. The lower pathway illustrates the progression from model
output and uncertainty assessment to evidence retrieval or abstention,
clinician review, human decision-making, and post-deployment monitoring.
Together, these components emphasize that predictive accuracy does not
necessarily confer clinical utility, technical capability does not constitute
clinical evidence, and automation does not imply deployment readiness. Strong
performance in any single dimension is therefore insufficient to establish
overall clinical safety, effectiveness, or readiness for clinical use.}
    \label{fig:evaluation_translation_pathway}
\end{figure}

These evaluation dimensions are integrated within a staged pathway from
benchmark assessment to prospective clinical validation, as presented in Figure~\ref{fig:evaluation_translation_pathway}.

\section{Discussion}

\subsection{Main Findings}

This review synthesizes medical Agentic AI across its conceptual boundaries,
core capabilities, technical architectures, application domains, evaluation
strategies, and requirements for clinical translation. The field is shifting
from task-specific predictive models toward goal-directed systems that
coordinate reasoning, tools, external knowledge, multimodal inputs, and
intermediate outputs across multistep workflows. Accordingly, medical Agentic
AI is best understood as a system-level orchestration framework rather than a
single model class.

Four recurrent architectural patterns emerged: single-agent tool-use systems,
multi-agent collaboration systems, knowledge-augmented agents, and multimodal
medical agents. These categories are not mutually exclusive; a single workflow
may combine foundation-model orchestration, specialized analytical tools,
external evidence retrieval, multimodal data integration, and clinician
feedback. System reliability therefore depends on the entire execution chain,
including tool selection, component performance, evidence quality, interaction
design, and error handling.

Applications now extend from medical question answering and text-based
reasoning to image interpretation, report generation, diagnostic assistance,
drug safety, clinical-trial analysis, and multimodal task coordination.
Medical imaging is particularly well suited to agentic approaches because it
requires the integration of visual findings, lesion localization, clinical
context, and textual interpretation. However, the current evidence base
remains dominated by prototype systems, public benchmarks, simulated
environments, retrospective datasets, and limited expert evaluation.
Existing studies therefore provide stronger evidence of technical feasibility
than of prospective clinical effectiveness, workflow benefit, or deployment
readiness.

\subsection{Comparison with Existing Reviews}


Previous reviews have approached healthcare agents from different conceptual
positions. Collaco et al. applied a relatively restrictive definition of
Agentic AI, emphasizing autonomous and goal-directed operation, and identified
a small number of eligible healthcare studies across emergency medicine,
oncology, radiology, and rehabilitation. Most were exploratory,
computational, or early-stage evaluations with limited real-world validation
\cite{Collaco2026AgenticHealthcare}. Zhao et al. adopted a broader AI-agent
perspective and reviewed applications in diagnostic assistance, clinical
decision support, report generation, patient interaction, hospital
management, medical education, and medication management
\cite{Zhao2026AIAgentHealthcare}.

Other surveys and perspectives approach the field from different angles (Table~\ref{tab:prior_reviews}). General surveys of LLM-based agents and multi-agent systems offer technical taxonomies but limited discussion of clinical validation \cite{Wang2024LLMAgentsSurvey,Xi2025RiseOfAgents,Guo2024MultiAgentsSurvey}. Medicine-focused perspectives emphasize opportunities for agentic systems in clinical care, biomedical discovery, and health-system coordination, together with the need for simulation-based evaluation and coordinated oversight \cite{Qiu2024AgenticSystemsMedicine,Gao2024BiomedicalAgents,Mehandru2024AgentsInClinic,Moritz2025CoordinatedAgents,Zou2025AgenticTeammates}. A recent survey of LLM-based medical agents organizes the field by architecture, application, and evaluation \cite{Wang2025BaymaxSurvey}, and a taxonomy-oriented review proposes multiple dimensions for characterizing agentic systems empirically \cite{Vatsal2026AgenticTaxonomy}. These works consistently identify a gap between technical capability and clinical evidence.

The present review extends this literature by linking conceptual definition,
functional capability, architectural organization, medical application,
evaluation, and clinical translation within a common analytical framework. It
also distinguishes complete agentic workflows from enabling components such
as vision--language models, segmentation systems, retrieval modules, and
general-purpose reasoning models. This distinction is important because the
presence of an advanced foundation model or isolated tool-use capability does
not, by itself, establish goal-directed agency, workflow autonomy, or clinical
readiness. The evidence-mapping approach further highlights the separation
between functional complexity and the maturity of reported clinical
validation.

\subsection{Limitations of Current Research}

A central limitation of the field is the absence of a stable and consistently
applied definition of Agentic AI. Terms such as ``agent'', ``AI agent'',
``LLM-based agent'', and ``Agentic AI'' are often used for systems with
substantially different capabilities. Some systems perform a single tool call,
whereas others maintain memory, revise actions in response to feedback, or
coordinate multiple agents over extended workflows. This terminological
variation complicates study comparison and may lead to the classification of
conventional pipelines or model-level reasoning systems as agentic systems.
Future studies should specify the system objective, planning mechanism,
available actions, tool interfaces, memory, feedback loops, degree of
autonomy, and conditions for human intervention.

The maturity of the evidence also remains limited. Most studies rely on
curated public datasets, retrospective records, simulated clinical
environments, or small-scale expert assessment. These settings provide useful
tests of technical capability but do not reproduce the incomplete data,
workflow interruptions, institutional variation, time pressure, and
accountability requirements of routine care. Strong performance on static
benchmarks should therefore not be interpreted as evidence of clinical
effectiveness. External validation, prospective silent deployment,
human--AI workflow studies, and controlled evaluations against current
clinical practice remain uncommon.

System-level error propagation represents a further concern. Agentic systems
combine multiple models, tools, retrieved sources, and decision steps, and an
error introduced early in the workflow may influence all subsequent outputs.
An inappropriate tool selection, incomplete retrieval result, incorrect code
execution, or inaccurate image localization may still produce a fluent and
apparently coherent final response. Evaluation must therefore examine
observable actions, intermediate outputs, evidence provenance, and recovery
from failure rather than final-answer accuracy alone.

Medical imaging introduces additional challenges in visual grounding and
cross-modal consistency. Generated interpretations should be traceable to
identifiable image findings, and textual claims should remain consistent with
detection, localization, or segmentation results. This requirement becomes
more demanding for volumetric imaging, longitudinal comparison, and
multimodal studies involving CT, MRI, PET, pathology, or combined image and
clinical data. Differences in acquisition protocols, devices, image quality,
annotation practices, and reporting conventions may further reduce
generalizability across institutions.

\subsection{Limitations of This Review}

This review has several limitations. First, the search was restricted to
English-language records published from 2022 onward and may therefore have
excluded earlier work described using different terminology or studies
published in other languages. The field is also developing rapidly, and
recent preprints, revised versions, or newly deployed systems may not have
been captured by the search date.

Second, Europe PMC records could not be exported at the individual-record
level and were retained only in the identification-stage count. These records
could not be independently deduplicated or screened using the same procedure
as records from the four exportable sources. This limitation may have affected
the completeness and precision of study identification and should be
considered when interpreting the reported selection totals.

Third, substantial heterogeneity in system definitions, study designs,
application domains, datasets, and evaluation settings precluded quantitative
meta-analysis and the use of a single conventional risk-of-bias instrument.
The evidence synthesis was therefore descriptive. The agentic complexity
score and validation-maturity categories were developed for comparative
mapping and should not be interpreted as validated measurements of system
quality, autonomy, clinical effectiveness, or deployment readiness.
Classification also depended on the completeness of reporting in the original
studies, and under-reporting of system components or validation procedures may
have influenced the assigned categories.

\subsection{Clinical Translation, Safety, and Real-World Validation}

Clinical translation requires substantially more than transferring a system
from a benchmark dataset to a hospital environment. Clinical data are
heterogeneous, incomplete, temporally evolving, and distributed across
multiple information systems. Relevant decisions may depend on imaging,
laboratory measurements, medication histories, prior interventions,
longitudinal records, and patient-reported information. Agentic systems must
therefore operate under missing information, inconsistent documentation, and
workflow constraints while preserving appropriate escalation to clinical
staff.

Safety concerns extend across the complete execution pathway. Responsible
medical AI requires attention to data quality, subgroup performance,
transparency, prospective monitoring, and mechanisms for preventing patient
harm \cite{Wiens2019DoNoHarm}. In agentic workflows, additional risks arise
from inappropriate tool invocation, unsupported retrieval, code-execution
errors, omission of clinically important information, propagation of
incorrect intermediate outputs, and failure to abstain or escalate. Safety
assessment should consequently address not only model predictions but also
action permissions, tool-level safeguards, uncertainty handling, fallback
procedures, and human oversight.

Reporting guidance should be selected according to study design and stage of
clinical evaluation. CONSORT-AI and SPIRIT-AI apply to reports and protocols
of clinical trials involving AI interventions
\cite{Liu2020CONSORTAI,Rivera2020SPIRITAI}. DECIDE-AI addresses early-stage
clinical evaluation, including human--AI interaction, workflow integration,
and safety \cite{Vasey2022DECIDEAI}. CLAIM provides imaging-specific guidance
for data, model development, validation, and clinical interpretation
\cite{Mongan2020CLAIM,Tejani2024CLAIM}. STARD-AI is intended for diagnostic
accuracy studies involving AI, whereas TRIPOD+AI applies to the development
and evaluation of clinical prediction models
\cite{Sounderajah2025STARDAI,Collins2024TRIPODAI}. FUTURE-AI provides broader
consensus principles for trustworthy and deployable healthcare AI, including
fairness, universality, traceability, usability, robustness, and
explainability \cite{Lekadir2025FUTUREAI}. The appropriate framework should
therefore be matched to the intended use, study design, and stage of
translation rather than applied as a generic checklist
\cite{Kolbinger2024ReportingGuidelinesAI}. Two recent developments are particularly relevant to agentic systems. TRIPOD-LLM extends reporting guidance to studies that develop, tune, or evaluate LLMs \cite{Gallifant2025TRIPODLLM}, and the CHART statement addresses studies that evaluate chatbot health advice \cite{CHART2025Statement}. For evidence synthesis, a PRISMA extension for AI has been proposed to improve the reporting of systematic reviews of AI studies \cite{Cacciamani2023PRISMAAI}, and the development of STARD-AI was preceded by a published protocol describing its consensus process \cite{Sounderajah2021STARDAIProtocol}. None of these guidelines yet fully addresses agent-specific elements such as tool inventories, action permissions, execution traces, memory, and inter-agent communication. Table~\ref{tab:reporting} maps the available guidelines to study types and lists the agent-specific reporting items recommended in this review.

\begin{table}[!htbp]
\centering
\caption{Reporting guidelines relevant to medical Agentic AI and recommended agent-specific reporting items. Guidelines should be selected according to study design and stage of translation; the agent-specific items proposed in this review are intended for use alongside, not instead of, existing checklists.}
\label{tab:reporting}
\footnotesize
\renewcommand{\arraystretch}{1.15}
\begin{tabularx}{\textwidth}{
>{\raggedright\arraybackslash}p{3.0cm}
>{\raggedright\arraybackslash}p{3.1cm}
>{\raggedright\arraybackslash}p{3.6cm}
>{\raggedright\arraybackslash}X}
\toprule
Guideline & Study type or stage & Relevance to Agentic AI & Recommended agent-specific items \\
\midrule
SPIRIT-AI and CONSORT-AI \cite{Rivera2020SPIRITAI,Liu2020CONSORTAI} & Trial protocols and reports & Prospective trials of agent-supported care & Agent version, tool inventory, action permissions, and human-oversight procedure \\
DECIDE-AI \cite{Vasey2022DECIDEAI} & Early-stage live clinical evaluation & First deployments involving clinician interaction & Logs of actions, overrides, abstentions, and escalations \\
STARD-AI \cite{Sounderajah2025STARDAI} & Diagnostic accuracy studies & Diagnostic agents & Information-gathering pathway and handling of abstentions \\
TRIPOD+AI \cite{Collins2024TRIPODAI} & Prediction model studies & Risk models invoked as agent tools & Calibration of tool outputs within the workflow \\
TRIPOD-LLM \cite{Gallifant2025TRIPODLLM} & Studies developing or evaluating LLMs & Backbone models, prompts, and evaluation & Prompt and tool specifications; model versions and access dates \\
CLAIM \cite{Mongan2020CLAIM,Tejani2024CLAIM} & Medical imaging AI & Imaging agents and perception tools & Visual grounding and image--text consistency analyses \\
CHART \cite{CHART2025Statement} & Chatbot health advice studies & Patient-facing conversational agents & Escalation behavior and safety-netting advice \\
FUTURE-AI \cite{Lekadir2025FUTUREAI} & Trustworthy AI across the life cycle & Design, deployment, and monitoring & Traceability of tool calls and evidence; monitoring and rollback plans \\
PRISMA-ScR and PRISMA-AI \cite{Tricco2018PRISMAScR,Cacciamani2023PRISMAAI} & Evidence synthesis & Reviews of agent studies & Operational definition of agency used for eligibility \\
\bottomrule
\end{tabularx}
\end{table}

Bias and generalizability remain major barriers to deployment. Medical data
vary across populations, institutions, devices, acquisition protocols,
clinical practices, and documentation systems. Models developed within a
single dataset or institution may therefore fail when transferred to new
settings. In imaging, differences in scanners, reconstruction methods, image
quality, and reporting conventions can affect both visual and textual
components. In EHR-based systems, changes in data schemas, coding practices,
and temporal documentation may disrupt tool use and reasoning. Prior research
has shown that dataset bias, weak external validation, and methodological
limitations can substantially overstate the apparent clinical performance of
medical AI systems
\cite{Nagendran2020AIvsClinicians,Kelly2019ClinicalImpactAI,
Liu2019DLvsProfessionals,Roberts2021COVIDPitfalls,
Varoquaux2022MedicalImagingFailures}.

Agentic systems also introduce deployment risks associated with adaptive and
context-dependent behavior. Subsequent actions may vary according to dialogue
history, retrieved evidence, intermediate results, or communication between
agents. In multi-agent systems, errors may be repeated, amplified, or converted
into superficial consensus. Auditability therefore requires structured
records of task decomposition, tool invocation, retrieved evidence,
intermediate outputs, abstention decisions, clinician interventions, and final
actions. Such records should support review without relying on unverifiable
internal reasoning narratives. Table~\ref{tab:failure_modes} summarizes the principal failure modes of medical Agentic AI across the execution pathway, together with representative evidence and mitigation strategies.

\begin{table}[!htbp]
\centering
\caption{Principal failure modes of medical Agentic AI, the stage of the workflow at which they arise, representative evidence, and mitigation strategies. Several failure modes interact; for example, an injected instruction may trigger inappropriate tool use whose output is then accepted through automation bias.}
\label{tab:failure_modes}
\footnotesize
\renewcommand{\arraystretch}{1.15}
\begin{tabularx}{\textwidth}{
>{\raggedright\arraybackslash}p{3.0cm}
>{\raggedright\arraybackslash}p{2.6cm}
>{\raggedright\arraybackslash}p{3.4cm}
>{\raggedright\arraybackslash}X}
\toprule
Failure mode & Where it arises & Representative evidence & Mitigation \\
\midrule
Hallucination and omission & Generation and summarization & \cite{Asgari2025HallucinationFramework,Kim2025MedicalHallucination,Ji2023HallucinationSurvey} & Evidence grounding; claim-level verification; omission checklists \\
Inappropriate tool selection or misuse & Planning and tool invocation & \cite{Ruan2024ToolEmu,Li2024MMedAgent} & Typed tool schemas; sandboxed execution; process-level evaluation \\
Irrelevant or outdated evidence & Retrieval & \cite{Xiong2024MedRAG,Zakka2024Almanac} & Curated sources; recency checks; citation verification \\
Code-execution errors & Data analysis & \cite{Shi2024EHRAgent,TayebiArasteh2024AutoML} & Sandboxes; unit tests; independent recomputation \\
Error propagation and false consensus & Multi-agent communication & \cite{Zhu2025MedAgentBoard,Du2024MultiagentDebate} & Independent evidence sources; disagreement reporting; human arbitration \\
Prompt injection and tool compromise & Inputs, retrieved content, and tools & \cite{Greshake2023IndirectInjection,Clusmann2025PromptInjection,Debenedetti2024AgentDojo,Hou2026MCP} & Input isolation; privilege separation; authenticated tools \\
Data poisoning and implanted misinformation & Training data and model parameters & \cite{Alber2025DataPoisoning,Han2024MisinformationAttacks} & Data provenance; integrity checks; knowledge-graph validation \\
Bias and inequity & Across the workflow & \cite{Omiye2023RaceBasedMedicine,Zack2024GPT4Bias} & Subgroup evaluation; fairness audits; representative data \\
Automation bias and over-reliance & Human--AI interaction & \cite{Goddard2012AutomationBias,Dratsch2023AutomationBiasMammography} & Uncertainty display; independent assessment; user training \\
Distribution shift and silent degradation & Deployment & \cite{Finlayson2021DatasetShift,Feng2022ClinicalAIQI} & Continuous monitoring; recalibration; rollback procedures \\
Privacy leakage and unauthorized actions & Memory, logging, and tool calls & \cite{Chan2024VisibilityAgents,Mandel2016SMARTonFHIR} & Least-privilege access; audit logs; data minimization \\
\bottomrule
\end{tabularx}
\end{table}

Simulated EHR environments, virtual hospitals, and multi-agent benchmarks
provide useful tests of interaction and tool use, but they cannot reproduce
the full complexity of clinical practice
\cite{Schmidgall2026AgentClinic,Jiang2025MedAgentBench,
Zhu2025MedAgentBoard,Fan2025AIHospital}. Real-world evaluation should proceed
through staged evidence generation, including external retrospective
validation, prospective silent testing, controlled workflow studies, and
post-deployment surveillance. Outcomes should include clinical performance,
workflow impact, subgroup reliability, clinician override, unintended
consequences, and patient-relevant effects rather than offline accuracy alone.

Human--AI collaboration requires explicit attention to how clinicians interact with agent outputs. Automation bias, the tendency to over-rely on automated advice, is well documented in clinical decision support \cite{Goddard2012AutomationBias}, and incorrect AI suggestions have been shown to degrade radiologists' performance in mammography reading \cite{Dratsch2023AutomationBiasMammography}. Randomized studies of LLM assistance have produced mixed effects on physician reasoning \cite{Goh2024DiagnosticReasoningRCT,Goh2025PatientCareRCT}, and LLM-drafted messages can shape the content of clinician responses \cite{Chen2024PatientMessages}. Agentic systems may intensify these effects because they present multistep conclusions with an appearance of procedural rigor. Interfaces should therefore expose evidence, uncertainty, and unresolved disagreement; workflows should preserve independent clinician assessment where appropriate; and studies should measure over-reliance and under-reliance in addition to accuracy.

Governance and regulation are evolving in parallel. Commentators have argued that LLM-based tools intended for medical purposes require approval as medical devices and that their safe use depends on regulators enforcing safety standards \cite{Gilbert2023ChatbotsAsDevices,Freyer2024RegulatorsSafety,Mesko2023RegulatoryOversight}. The US Food and Drug Administration has reviewed its regulation of AI-enabled products and highlighted concepts that merit consideration as the regulatory system adapts to the challenges posed by AI, including generative models \cite{Warraich2025FDAPerspective}. For agentic systems, governance must extend beyond model outputs to delegated actions, including which tools an agent may access, which actions require approval, how activity is logged and monitored, and who is accountable when an action causes harm \cite{Chan2024VisibilityAgents,Ning2024GenAIEthics}. Documentation practices such as model facts labels could be extended to describe tool inventories, permissions, known failure modes, and monitoring plans \cite{Sendak2020ModelFacts}. Because performance can drift as data, workflows, tools, and underlying models change, continual monitoring and predefined update procedures are required \cite{Finlayson2021DatasetShift,Feng2022ClinicalAIQI}.

\begin{figure}[!t]
    \centering
\includegraphics[
    width=\textwidth,
    trim=0.2cm 0.5cm 0.2cm 0cm,
    clip
]{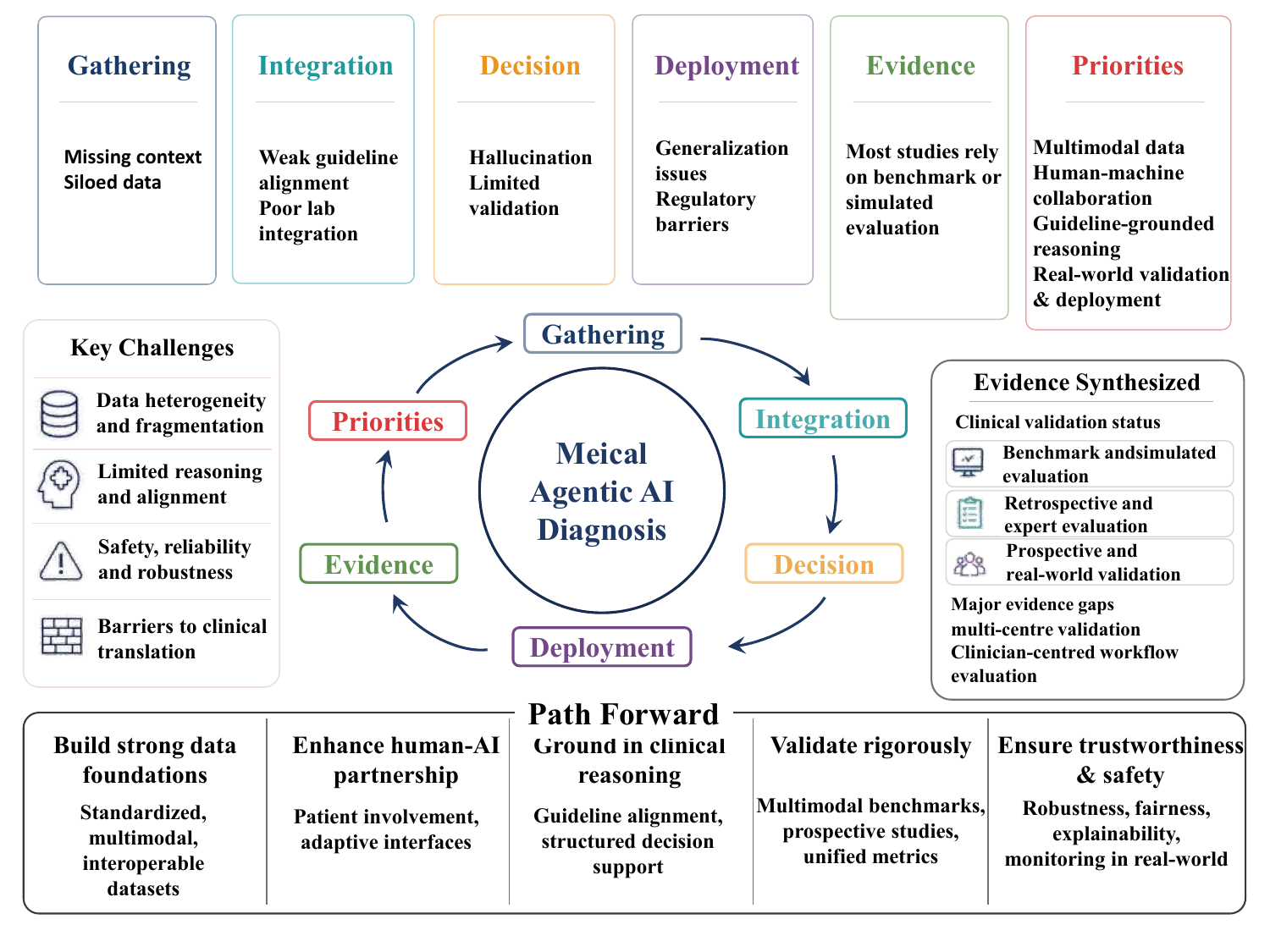}
\caption{Current challenges and research priorities for medical
Agentic AI.
Major barriers span multimodal data integration, planning and
reasoning, tool and evidence reliability, uncertainty and safety, clinical
validation, and real-world deployment. The evidence base remains dominated by
public benchmarks, simulated environments, retrospective datasets, and limited
expert evaluation. Key priorities include interoperable multimodal data,
traceable tool use, evidence-grounded and uncertainty-aware decision processes,
effective human--AI collaboration, external and prospective validation, and
post-deployment monitoring. These priorities define a research and translation
agenda and should not be interpreted as evidence that current systems are ready
for trustworthy clinical deployment.}
    \label{fig:medical_agentic_ai_diagnosis}
\end{figure}

\subsection{Future Directions}

Future research should prioritize clinically grounded evaluation rather than
continued expansion of benchmark capability alone. Benchmarks should represent
multistep tasks, incomplete information, multimodal inputs, realistic tool
interfaces, and clinically meaningful consequences of error. Evaluation
should include external datasets, multi-institutional studies, subgroup
analysis, prospective silent deployment, and human--AI workflow assessment.
Calibration, uncertainty-aware abstention, escalation behavior, revision
burden, response latency, and effects on clinical decisions should be reported
alongside conventional task metrics.

Technical development should emphasize modular reliability and evidence
traceability. Foundation models, specialized analytical tools, retrieval
systems, memory components, and human review interfaces should be evaluated
both independently and as an integrated workflow. Systems should record which
tools were called, which evidence was retrieved, how intermediate outputs were
used, and where uncertainty entered the process. In medical imaging, priority
areas include visual grounding, cross-modal consistency, volumetric and
longitudinal reasoning, robust localization, and validation across devices and
institutions. Segmentation, registration, image fusion, and report-generation
modules may be coordinated within agentic workflows, but their clinical value
must be established through downstream outcomes rather than visual quality or
workflow complexity alone.

Governance must develop in parallel with technical capability. Future studies
should define action permissions, responsibility boundaries, clinician
override mechanisms, privacy protections, update procedures, and criteria for
suspending or withdrawing a system. Adaptive agents may change their behavior
after model updates, knowledge-base revisions, or changes in available tools,
making continuous monitoring essential. Post-deployment evaluation should
therefore examine distribution shift, tool failure, emerging safety signals,
subgroup disparities, user adaptation, and unintended workflow effects.

The evidence reviewed points to five specific priorities. First, agent-specific training and evaluation environments should move from static questions toward realistic and verifiable tasks. Environments that support both the training and evaluation of agents, such as virtual EHRs and executable biomedical task suites, may enable learning from task outcomes while preserving auditability \cite{Jiang2025MedAgentBench,Xu2025MedAgentGym}. Second, evaluations should report cost, latency, and resource use alongside accuracy, because multistep and multi-agent systems can be substantially more expensive than single-model baselines and because cost-aware decision-making is itself clinically relevant \cite{Kapoor2025AgentsThatMatter,Nori2026SequentialDiagnosis}. Third, standardized and secure interfaces to clinical systems, built on FHIR-based access, authenticated tool registries, and least-privilege permissions, are needed before agents can be connected safely to operational EHRs \cite{Mandel2016SMARTonFHIR,Hou2026MCP}.

Fourth, prospective evidence should be generated in stages, beginning with silent deployment and progressing to controlled workflow studies and pragmatic trials with patient-relevant outcomes. Evaluations of language models within health systems and large-scale deployments of ambient documentation tools indicate that such studies are feasible in operational settings \cite{Jiang2023NYUTron,Tierney2024AmbientScribes}. Finally, human--AI teaming should be studied as a design problem in its own right, with attention to how agents communicate uncertainty, request clarification, defer to clinicians, and support rather than displace clinical judgment \cite{Zou2025AgenticTeammates,Moritz2025CoordinatedAgents}. Table~\ref{tab:research_agenda} links these priorities to the challenges identified in this review and to the evidence required to address them.

\begin{table}[!htbp]
\centering
\caption{Research agenda for medical Agentic AI. The five priorities correspond to those shown in Figure~\ref{fig:medical_agentic_ai_diagnosis}; for each, the table summarizes the current gap, representative research directions, and the evidence needed to demonstrate progress.}
\label{tab:research_agenda}
\footnotesize
\renewcommand{\arraystretch}{1.15}
\begin{tabularx}{\textwidth}{
>{\raggedright\arraybackslash}p{2.7cm}
>{\raggedright\arraybackslash}p{3.4cm}
>{\raggedright\arraybackslash}X
>{\raggedright\arraybackslash}p{3.6cm}}
\toprule
Priority & Current gap & Research directions & Evidence needed \\
\midrule
Build strong data foundations & Siloed, heterogeneous, and incomplete multimodal data & Standardized, multimodal, and interoperable datasets; FHIR-based access; linked imaging, EHR, and genomic records \cite{Mandel2016SMARTonFHIR,Johnson2023MIMICIV} & Multi-institutional datasets with documented provenance \\
\addlinespace
Enhance human--AI partnership & Automation bias; limited patient involvement; static interfaces & Adaptive interfaces that expose evidence and uncertainty; patient involvement in design \cite{Goddard2012AutomationBias,Zou2025AgenticTeammates} & Human--AI collaboration studies measuring reliance, workload, and decision quality \\
\addlinespace
Ground in clinical reasoning & Weak guideline alignment; poor integration of laboratory and contextual data & Guideline-grounded planning; structured decision support; verifiable tool use \cite{Hager2024ClinicalDecisionLimits,Wang2025MedAgentPro} & Guideline concordance and process-level evaluation \\
\addlinespace
Validate rigorously & Predominance of benchmark and simulated evaluation & Multimodal and interactive benchmarks; unified process metrics; prospective studies \cite{Bedi2025TestingLLMs,Nori2026SequentialDiagnosis} & External, prospective, and randomized evaluations with patient-relevant outcomes \\
\addlinespace
Ensure trustworthiness and safety & Hallucination, adversarial vulnerability, bias, and limited monitoring & Robustness and security testing; fairness audits; explainability; post-deployment monitoring \cite{Alber2025DataPoisoning,Omiye2023RaceBasedMedicine,Feng2022ClinicalAIQI} & Safety-event reporting; subgroup analyses; drift monitoring \\
\bottomrule
\end{tabularx}
\end{table}

The principal limitations of current medical Agentic AI and the corresponding
research priorities are summarized in
Figure~\ref{fig:medical_agentic_ai_diagnosis}.

\section{Conclusion}

Medical Agentic AI has become an important direction in the recent development of medical artificial intelligence. Unlike traditional medical AI systems that mainly focus on single tasks such as classification, detection, segmentation, or text generation, Agentic AI emphasizes task decomposition, tool use, external knowledge retrieval, multi-step reasoning, and result integration around specific medical goals. Existing studies suggest that medical Agentic AI has shown potential in medical image interpretation, radiology report generation, multimodal medical image analysis, diagnostic assistance, clinical reasoning, electronic health record analysis, treatment planning, drug safety, clinical trial analysis, and biomedical research. Among these areas, medical image-related tasks are particularly suitable for demonstrating the value of agentic systems in multimodal information integration, downstream task coordination, and result interpretation.

However, medical Agentic AI is still at an early stage of development. The conceptual boundaries among AI agents, LLM-based agents, and Agentic AI remain inconsistent across studies, and system architectures and implementation strategies also vary substantially. More importantly, most existing studies still rely on public datasets, simulated environments, retrospective experiments, or small-scale expert evaluation. Systems that have undergone multi-center, prospective, and real clinical workflow validation remain limited. Therefore, current medical Agentic AI should be regarded more appropriately as an emerging technical paradigm rather than a mature clinical solution.

Future research should move beyond the sole pursuit of model performance and pay more attention to system validation in real medical scenarios. For medical imaging, future studies should not only focus on improving metrics for individual tasks such as classification, detection, segmentation, or report generation, but also strengthen image evidence localization, cross-modal consistency validation, result interpretability, and physician-reviewable mechanisms. At the same time, evaluation standards for medical Agentic AI need to be further improved. In addition to traditional performance metrics, future evaluations should consider task completion rate, tool-use reliability, evidence traceability, safety, clinical usability, and compatibility with real clinical workflows. Only when image analysis performance, multimodal information integration, result interpretability, and clinical validation are sufficiently supported can medical Agentic AI become a reliable intelligent system for assisting medical image understanding and clinical decision-making.

\newpage
\section*{Declaration statements}

\subsection*{Funding}
This work was supported in part by the National Natural Science Foundation of China (No. 62502064), Joint Plan of Liaoning Province Science and Technology Plan (No. 2025JH2/101800417), Scientific Research Project of Liaoning Provincial Department of Education (No. LJ222511258003), Joint Plan of Liaoning Province Science and Technology Plan (No. 2025JH2/101800422), Interdisciplinary Project of Dalian University (No. DLUXK-2025-QN-020), 111 Center (No. D23006).

\subsection*{Conflict of interest/Competing interests}
The authors declare no competing interests.

\subsection*{Ethics approval and consent to participate}
Not applicable. This review does not involve new experiments on humans, animals, or human samples.

\subsection*{Consent for publication}
Not applicable.

\subsection*{Data availability}
No new datasets were generated or analyzed in this review.

\subsection*{Materials availability}
Not applicable.

\subsection*{Code availability}
Not applicable.

\newpage
\subsection*{Author Contributions}

Wanshu Fan conceptualized the review, defined its scope, and developed the overall analytical framework. Wanshu Fan contributed to the review methodology, evidence-synthesis strategy, manuscript organization, refinement of conceptual definitions and terminology, interpretation of the evidence, and revision of the manuscript.

Zheng Tong conducted the literature search, study screening, evidence organization, and synthesis, and was primarily responsible for drafting the initial manuscript. Zheng Tong also contributed to technical classification, reference verification, integration of different sections, and subsequent manuscript revision.

Yang Liu contributed to literature collection, evidence organization, and verification of relevant studies. Yang Liu was responsible for the conception, design, preparation, and revision of all figures, and ensured consistency of terminology, data, classifications, and figure captions with the main text throughout manuscript revision.

Zhongbin Han reviewed the manuscript from a clinical perspective and provided expert input on medical terminology, diagnostic and treatment workflows, medical image interpretation, clinical decision support, patient safety, clinical usability, and real-world clinical translation.

Haifan Gong, Congyu Liao, and Xiaofeng Liu provided expert guidance on medical artificial intelligence and related technical aspects. They reviewed the conceptual definitions, technical architectures, representative studies, medical applications, evaluation approaches, and clinical translation challenges, and provided comments on the technical content and academic interpretation.

Jing Qin contributed to the review methodology, analytical framework, and interpretation of the findings. Jing Qin critically reviewed and revised the manuscript, with particular attention to methodological rigor, technical accuracy, logical consistency, and academic presentation.

Cong Wang supervised the study and provided strategic guidance on the research direction, review framework, and manuscript organization. Cong Wang critically revised the manuscript for important intellectual content and oversaw its final refinement and preparation for submission.

All authors reviewed and approved the final manuscript.

\newpage
\bibliography{references}

\end{document}